\documentclass[letterpaper]{article}

\usepackage{uai2018}
\usepackage[margin=1in]{geometry}
 \usepackage{times}

\usepackage[utf8]{inputenc}
\usepackage[english]{babel}
\usepackage{tikz, pgf}
\usepackage{natbib}
\usepackage{hyperref}

\usepackage{booktabs}
\usepackage{multirow, bigstrut}
\usepackage{algorithm}
\usepackage{algpseudocode}
\usepackage{epsfig,amssymb,amsmath,ifthen,comment}
\usepackage{float}
\usepackage{xcolor}
\usepackage{xr-hyper}
\usepackage{mathtools}
\usepackage{subcaption}
\usepackage{caption}

\usepackage{xr}

\usepackage{diagbox}

\DeclareMathOperator*{\argmax}{arg\,max}

\newcommand{\E}{\mathbb{E}}

\DeclareMathOperator{\pa}{pa}
\DeclareMathOperator{\ch}{ch}

\newtheorem{thm}{Theorem}

\newenvironment{thma}[1]{\par\noindent{\bf Theorem #1\ }\em}{\em}

\newenvironment{prf}{\noindent\textit{Proof:}\begin{mdseries}}{\end{mdseries}{\hfill\scriptsize$\Box$}} 

\title{Estimation of Personalized Effects Associated With Causal Pathways}

\author{} 

\author{ {\bf Razieh  Nabi} \\
Computer Science Dept.  \\
Johns Hopkins University\\
Baltimore, MD, 21218 \\
\And
{\bf Phyllis Kanki} \\
Immunology and Infectious Diseases Dept. \\
Harvard T. H. Chan School of Public Health \\
Boston, MA, 02115 \\
\And
{\bf Ilya Shpitser}  \\
Computer Science Dept.  \\
Johns Hopkins University\\
Baltimore, MD, 21218 \\
}

\begin{document}

\maketitle

\begin{abstract}
The goal of personalized decision making is to map a unit's characteristics to an action tailored to maximize the expected outcome for that unit.  Obtaining high-quality mappings of this type is the goal of the dynamic regime literature.
In healthcare settings, optimizing policies with respect to a particular causal pathway may be of interest as well.  For example, we may wish to maximize the chemical effect of a drug given data from an observational study where the chemical effect of the drug on the outcome is entangled with the indirect effect mediated by differential adherence.  In such cases, we may wish to optimize the direct effect of a drug, while keeping the indirect effect to that of some reference treatment. 
\citep{Shpitser18IDpolicies} shows how to combine mediation analysis and dynamic treatment regime ideas to defines policies associated with causal pathways and counterfactual responses to these policies.
In this paper, we derive a variety of methods for learning high quality policies of this type from data, in a causal model corresponding to a longitudinal setting of practical importance.  We illustrate our methods via a dataset of HIV patients undergoing therapy, gathered in the Nigerian PEPFAR program.
\end{abstract}

\section{INTRODUCTION}
\label{sec:intro}

There has been growing interest in making personalized decisions in different domains to account for inherent heterogeneity among 
individuals and optimize individual-level experiences. 
For instance, personalized medicine aims at systematic use of individual patient history 
including biological information and biomarkers to improve patient's health care. Personalized actions can be viewed as realizations of decision rules 
where available information is mapped to the space of possible decisions.  

Making good personalized decisions
often involves acting in multiple stages. 
For instance, multiple successive medical interventions may be required for long-term care of patients with chronic diseases.
The goal of personalized medicine is to tailor a sequence
of decision rules on treatment, known as dynamic treatment regimes, based on patient characteristics seen so far, to maximize the likelihood of a desirable outcome.
A number of algorithms have been developed for finding optimal treatment regimes from either observational data, or data from experiments tailored for providing information on regime quality, such as sequential multiple assignment randomized trials (SMARTs) \citep{Lavori00SMART, Murphy05SMART}.
These algorithms use methods from causal inference, and aim to predict counterfactual outcomes under policies different from those actually followed in the data \citep{robins04SNMM, murphy03optimal, Moodie13DTR}.

A natural extension of these methods is finding treatment regimes that optimizes a \emph{part} of the effect of the treatment on the outcome.  We illustrate the utility of this problem with the following example.  Patients infected with human immunodeficiency virus (HIV) are typically put on courses of antiretroviral therapy, as a first line of therapy. Although these sort of medications are effective in combating the disease, the full benefit is not realized since patients often do not fully adhere to the medication regimen. One of the causes of poor adherence to the therapy is toxicity of the medication.  Hence, viral failure in a patient receiving treatment may be attributed to either poor drug effectiveness or lack of adherence to an otherwise effective treatment plan.  In observational studies of HIV patients, treatments are not randomly assigned, and patients have differential adherence.  Because of this, finding a policy that optimizes the \emph{overall effect} of the treatment plan on the outcome entangles two very different causal pathways -- the chemical pathway associated with the active ingredients in the treatment, and the pathway associated with adherence.

We may, instead, consider the problem of finding a set of policies that  optimize only the direct chemical effect of the drug, in the counterfactual situations where the indirect effect mediated by adherence can be kept to that of some reference treatment.  Policies of this type may be more directly relevant in precision medicine contexts where adherence varies among patients.

\citep{Shpitser18IDpolicies} defines counterfactual responses to policies that set treatments only with respect to a particular causal pathway, and gives a general identification algorithm for these responses, as a generalization of similar algorithms for standard dynamic treatment regimes \citep{tian02onid}, and effects associated with causal pathways in mediation analysis \citep{shpitser13cogsci}.  In this paper, we consider algorithms that can be used to find policies that maximize effects along particular causal pathways from data, in a causal model of most immediate relevance in longitudinal settings.

The paper is organized as follows. We fix our notation, and define counterfactual responses to treatment policies and policies associated with pathways in Section \ref{sec:notations}.  In Section \ref{sec:opt-overall-policy}, we review existing techniques used in learning policies optimizing the \emph{overall effect} of actions.  In Section \ref{sec:opt-edge-policy}, we fix the causal model corresponding to a typical longitudinal study, prove identification of counterfactual policy responses under this model, and show how techniques for maximizing treatment policies, described in Section \ref{sec:opt-edge-policy},
may be extended to maximize policies associated with causal pathways.  In Section \ref{sec:experiment}, we illustrate the methods we propose via an application involving data on treatment of HIV patients in Nigeria.
Our conclusions are in Section \ref{sec:conclusion}.
The Appendix contains the proofs of all claims, a basic description of statistical inference in semi-parametric models, as context for some of our estimation strategies, visualizations of learned policies, and descriptions of the experiments.

\section{NOTATIONS AND PRELIMINARIES}
\label{sec:notations}
Consider a multi-stage decision problem with $K$ pre-specified decision points, indexed by $i = 1, \ldots, K$. Let $Y$ denote the final outcome of interest and $A_i$ denote the action made at decision point $i$ with the finite state space of ${\mathfrak X}_{A_i}$.
The set of all actions is denoted by $\bf{A}$. 
Let $W_0$ denote the available information prior to the first decision, and $W_i$ denote the information collected between decisions $i$ and $i+1$, ($Y \equiv W_{K}$). 
Given $A_i$, denote $\overline{A}_i$ to be all treatments administered from time $1$ to $i$, 
similarly for $W_i$ and $\overline{W}_i$. 
We combine the treatment and covariate history up to treatment decision $A_i$ into a history vector $H_i$.
The state space of $H_i$ is denoted by ${\mathfrak X}_{{H}_i}$. 

We are interested in learning policies that map $H_i$ to values of $A_i$, for all $i$, that maximize the expected value of the outcome $Y$.  Doing this from observed data entails considering \emph{counterfactual outcomes} $Y$ had $A_i$ been assigned in a different way from what was actually observed.  We briefly review graphical causal models, and the potential outcome notation from causal inference, which will be used to define such counterfactuals.

Causal models are sets of distributions defined by restrictions associated with directed acyclic graphs (DAGs).
We will use vertices and variables interchangeably -- capital letters for a vertex or variable ($V$), bold capital letter for a set (${\bf V}$), lowercase letters for values ($v$), and bold lowercase letters for sets of values (${\bf v})$. By convention, each graph is defined on a vertex set $\mathbf{V}$.
For a set of values ${\bf a}$ of ${\bf A}$, and a subset ${\bf A}^{\dag} \subseteq {\bf A}$, define ${\bf a}_{{\bf A}^{\dag}}$ to be a restriction of ${\bf a}$ to elements in ${\bf A}^{\dag}$.
For a DAG ${\cal G}$, and any $V \in {\bf V}$, we define the parents of $V \in {\bf V}$ to be the set
$\pa_{\cal G}(V) \equiv \{ W \in {\bf V} \mid W \to V \}$, and the children of $V \in {\bf V}$ to be the set
$\ch_{\cal G}(V) \equiv \{ W \in {\bf V} \mid V \to W \}$.


Causal models of a DAG ${\cal G}$ consist of distributions defined on counterfactual random variables
of the form $V({\bf a})$ where ${\bf a}$ are values of $\pa_{\cal G}(V)$.  These variables represent outcomes of $V$ had all variables in $\pa_{\cal G}(V)$ been set, possibly contrary to fact, to ${\bf a}$.
In this paper we assume Pearl's functional model for a DAG $\mathcal{G}$ with vertices $\mathbf{V}$ which is the set containing any joint distribution over all potential outcome random variables where the sets of variables 
$\big\{ \{V(\mathbf{a}_V) \mid \mathbf{a}_V \in \mathfrak{X}_{\pa_{\mathcal{G}}(V)} \} \mid V \in \mathbf{V} \big\}$
are mutually independent \citep{pearl09causality}.
The \emph{atomic counterfactuals} in the above set model the relationship between $\pa_{\cal G}(V)$, representing direct causes of $V$, and $V$ itself.  From these, all other counterfactuals may be defined using \emph{recursive substitution}.  For any ${\bf A} \subseteq {\bf V} \setminus \{ V \}$,
{\small
\begin{align}
V({\bf a}) \equiv V({\bf a}_{\pa_{\cal G}(V) \cap {\bf A}},
\{ \pa_{\cal G}(V) \setminus {\bf A} \}({\bf a})),
\label{eqn:rec-sub}
\end{align}
}
where $\{ \pa_{\cal G}(V) \setminus {\bf A} \}({\bf a}))$ is taken to mean the (recursively defined) set of counterfactuals associated with variables in $\pa_{\cal G}(V)$, had ${\bf A}$ been set to ${\bf a}$.  

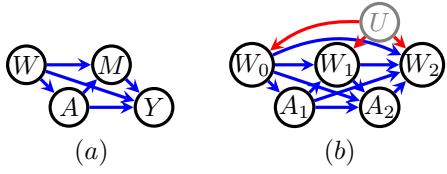
\begin{figure}
\begin{center}
\begin{tikzpicture}[>=stealth, node distance=0.8cm]
    \tikzstyle{format} = [draw, very thick, circle, minimum size=5.0mm,
	inner sep=0pt]
	\begin{scope}
		\path[->, very thick]
			node[format] (w) {$W$}
			node[format, below right of=w] (a) {$A$}
			node[format, above right of=a] (m) {$M$}
			node[format, below right of=m] (y) {$Y$}
			(w) edge[blue] (a)
			(a) edge[blue] (y)
			(a) edge[blue] (m)
			(m) edge[blue] (y)
			(w) edge[blue] (m)
			(w) edge[blue] (y)
			node[below of=a, yshift=0.2cm, xshift=0.3cm] (l) {$(a)$}	
			;
	\end{scope}
	
	\begin{scope}[xshift=3.0cm] 
		\path[->, very thick]
			node[format] (w) {$W_0$}
			node[format, below right of=w] (a) {$A_1$}
			node[format, above right of=a] (m) {$W_1$}
			node[format, gray, above right of=m] (u) {$U$}
			node[format, below right of=m] (z) {$A_2$}
			node[format, above right of=z] (y) {$W_2$}
			(u) edge[red, bend right=20] (w)
			(w) edge[blue] (a)
			(a) edge[blue] (z)
			(a) edge[blue] (m)
			(m) edge[blue] (z)
			(w) edge[blue] (m)
			(w) edge[blue] (z)
			(z) edge[blue] (y)
			(m) edge[blue] (y)
			(a) edge[blue] (y)
			(w) edge[blue, bend left=25] (y)
			(u) edge[red] (m)
			(u) edge[red] (y)
			node[below of=a, yshift=0.2cm, xshift=0.6cm] (l) {$(b)$}
		;
	\end{scope}
\end{tikzpicture}
\end{center}
\caption{(a) A simple causal DAG, with a single treatment $A$, a single outcome $Y$, a vector $W$ of baseline variables, and a single mediator $M$.
(b) A more complex causal DAG with two treatments $A_1,A_2$, an intermediate outcome $W_1$, and the final outcome $W_2$.
}
\label{fig:triangle}
\end{figure}

A causal parameter is said to be \emph{identified} in a causal model if it is a function of the observed data distribution $p({\bf V})$.
In all causal models of a DAG ${\cal G}$ in the literature, all interventional distributions 
$p(\{ {\bf V} \setminus {\bf A} \}({\bf a}))$
are identified by the \emph{g-formula}:
{\small
\begin{align}
p(\{ {\bf V} \setminus {\bf A} \}({\bf a})) =
\!\!\!
\prod_{V \in {\bf V} \setminus {\bf A}}
\left.
\!\!\!
p(V | \pa_{\cal G}(V)) \right|_{{\bf A}={\bf a}}
\label{eqn:g}
\end{align}
}\noindent
As an example, $Y(a)$ in the DAG in Fig.~\ref{fig:triangle} (a), is defined to be $Y(a,M(a,W),W)$, 
and its distribution is identified as $\sum_{w,m} p(Y | a,m,w) p(m | a,w) p(w)$.
In our sequential decision setting, the relevant counterfactual is
$Y(\overline{a}_K)$, i.e. the response of $Y$ had the treatment history $\overline{A}_K = \overline{a}_K$ been administered, possibly contrary to fact. Comparison of $Y(\overline{a}_K)$ and $Y(\overline{a}'_K)$ in expectation, $\E[Y(\overline{a}_K)] - \E[Y(\overline{a}'_K)]$, where $\overline{a}_K$ is the treatment history of interest, and $\overline{a}'_K$ is the reference treatment history,
gives the average causal effect of $\overline{a}_K$ on the outcome $Y$.

\subsubsection*{Counterfactual Response to Policies}  
A dynamic treatment regime ${\bf f_A} = \{f_{A_1}, \ldots, f_{A_K} \}$ is a sequence of decision rules that forms a treatment plan for patients over time.  At the $i$th decision point, the $i$th rule $f_{A_i}$ maps the available information $H_i$ prior to the $i$th decision point to a treatment $a_i$, i.e. $f_{A_i}: {\mathfrak X}_{{H}_i} \mapsto {\mathfrak X}_{{A}_i}$.
Given a treatment regime ${\bf f_A}$, we define the counterfactual response $Y$ had ${\bf A}$ been assigned according to ${\bf f}_{\bf A}$,
or $Y({\bf f_A})$, as the following generalization of (\ref{eqn:rec-sub})
{\scriptsize
\begin{align}
Y(
\{ f_{A_i}(H_i({\bf f}_{\bf A})) | A_i \in \pa_{\cal G}(Y)\cap {\bf A} \},
\{ \pa_{\cal G}(Y) \setminus {\bf A} \}({\bf f_A})
).
\label{eqn:rec-sub-f}
\end{align}
}
%
Under a causal model associated with the DAG $\cal G$, 
the distribution $p(Y({\bf f_A}))$, is identified as the following generalization of the g-formula
{\scriptsize
\begin{align}
\sum_{{\bf V} \setminus Y, {\bf A}} \! \prod_{V \in {\bf V} \setminus {\bf A}}
\!\!\!
\!\!\!
p
(
V |
\{ f_{A_i}(H_i) | A_i \in \pa_{\cal G}(V) \cap {\bf A} \},\!
\pa_{\cal G}(V) \setminus {\bf A} 
)
\label{eqn:g-f}
\end{align}
}
As an example, $Y(a = f_A(W))$ in Fig.~\ref{fig:triangle} (a) is defined as
$Y(a=f_A(W), M(a=f_A(W),W), W)$, 
%
and its distribution is identified as $\sum_{w,m} p(Y | a=f_A(w),m,w) p(m | a=f_A(w),w) p(w)$.
We will call policies corresponding to counterfactuals defined in (\ref{eqn:rec-sub-f}) \emph{overall policies}, to distinguish them from policies associated with causal pathways to be defined later.

\subsection*{Mediation 
Analysis}
An important goal of causal inference is understanding the mechanisms by which the treatment $A$ influences the outcome $Y$.  A common framework for mechanism analysis is \textit{mediation analysis} which seeks to decompose the effect of $A$ on $Y$ into the \textit{direct effect} and the \textit{indirect effect} mediated by a third variable, or more generally into components associated with particular causal pathways.  Consider the graph in Fig.~\ref{fig:triangle} (a): the direct effect corresponds to the effect along the edge $A \rightarrow Y$, and indirect effect corresponds to the effect along the path $A \rightarrow M \rightarrow Y$, mediated by $M$.

Counterfactuals associated with mediation analysis have been defined using a more general type of intervention in a graphical causal model,
namely the \emph{edge intervention} \citep{shpitser15hierarchy}, which maps a set of directed edges in ${\cal G}$ to values of their source vertices.  Edge interventions have a natural interpretation in cases where a treatment variable has multiple components that a) influence the outcome in different ways, b) occur or do not occur together in observed data, and  c) may in principle be intervened on separately.  For instance, smoking leads to poor health outcomes due to two components: smoke inhalation and exposure to nicotine.  A smoker would be exposed to both of these components, while a non-smoker to neither.  However, one might imagine exposing someone selectively only to nicotine but not smoke inhalation (via a nicotine patch), or only smoke inhalation but not nicotine (via smoking plant matter not derived from tobacco leaves).  These types of hypothetical experiments correspond precisely to edge interventions, and have been used to conceptualize direct and indirect effects \citep{robins92effects, pearl01direct}.

We will write the mapping of a set of edges to values of their source vertices as
${\mathfrak a}_{\alpha}$ to mean edges in $\alpha$ are mapped to values in the \emph{multiset} ${\mathfrak a}$ (since multiple edges may share the same source vertex, and be assigned to different values).  For a subset $\beta \subseteq \alpha$ and an assignment ${\mathfrak a}_\alpha$, denote ${\mathfrak a}_{\beta}$ to be a restriction of ${\mathfrak a}_{\alpha}$ to edges in $\beta$.  For simplicity, in the remainder of the paper we assume that if $(AW)_{\to} \in \alpha$, then for all $V \in \ch_{\cal G}(A)$, $(AV)_{\to} \in {\alpha}$, {where $(XY)_{\to}$ is a directed edge from X to Y}. 

We will write counterfactual responses to edge interventions as $Y({\mathfrak a}_{\alpha})$.
An edge intervention that sets a set of edges $\alpha$ to values in the multiset ${\mathfrak a}$
is defined via the following generalization of recursive substitution (\ref{eqn:rec-sub}):
{\small
\begin{align}
Y({\mathfrak a}_{\alpha}) \equiv Y({\mathfrak a}_{\{ (ZY)_{\to} \in \alpha \}},
\{ \pa^{\bar{\alpha}}_{\cal G}(Y) \}({\mathfrak a}_{\alpha})),
\label{eqn:rec-sub-e}
\end{align}
}
where 
$\pa^{\bar{\alpha}}_{\cal G}(Y) \equiv \{ W \mid (WY)_{\to} \not\in \alpha \}$.
For example, in the DAG in Fig.~\ref{fig:triangle} (a), $Y({\mathfrak a}_{\{(AY)_{\to},(AM)_{\to}\}})$ assigning $(AY)_{\to}$ to $a$
and $(AM)_{\to}$ to $a'$ is defined as $Y(a,M(a',W),W)$.

Identification of edge interventions in graphical causal models of a DAG corresponds quite closely with identification of regular (node) interventions, as follows.  Let ${\bf A}_{\alpha} \equiv \{ A \mid (AB)_{\to} \in \alpha \}$. Consider an edge intervention given by the mapping ${\mathfrak a}_{\alpha}$. Then, under the functional model of a DAG ${\cal G}$, the joint distribution of counterfactual responses
$p(\{  {\bf V} \setminus {\bf A}_{\alpha} \}({\mathfrak a}_{\alpha}))$
is identified via the following generalization of (\ref{eqn:g}) called the
\emph{edge g-formula}:
{\small
\begin{align}
\prod_{V \in {\bf V} \setminus {\bf A}_{\alpha}} p(V | {\mathfrak a}_{\{ (ZV)_{\to} \in \alpha \}}, \pa^{\bar{\alpha}}_{\cal G}(V) ).
\label{eqn:g-e}
\end{align}
}
For example, in Fig \ref{fig:triangle} (a), $p(Y({\mathfrak a}_{\{(AY)_{\to}, (AM)_{\to}\}})) = \sum_{W,M} p(Y | a, M, W) p(M | a', W) p(W)$, which is obtained by marginalizing $W$ and $M$ from the edge g-formula.

\subsubsection*{Counterfactual Responses To Policies Associated With Pathways}

We now define counterfactual responses to policies that operate only with respect to particular outgoing edges from ${\bf A}$.  These counterfactuals generalize those in the previous two sections.

As an example, we can view Fig.~\ref{fig:triangle} (a) as representing a cross-sectional study of HIV patients of the kind described in \citep{miles17quantifying}, where $W$ is a set of baseline characteristics, $A$ is one of a set of possible antiretroviral treatments, $M$ is adherence to treatment, and $Y$ is a binary outcome variable signifying viral failure.  In this type of study, we may wish to find $f_A(W)$
that maximizes the expected outcome $Y$ had $A$ been set according to $f_A(W)$ for the purposes of the direct effect of $A$ on $Y$,
and $A$ were set to some reference level $a'$ for the purposes of the effect of $A$ on $M$.  In other words, we may wish to find $f_A(W)$
to maximize the counterfactual mean $\E[Y(f_A(W), M(a',W), W)]$.  This would correspond to finding a treatment policy that maximizes the direct (chemical) effect, if it were possible to keep adherence to a level $M(a')$ as if a reference (easy to adhere to) treatment $a$ were given.

We now give a general definition for responses to such path-specific policies. Fix a set of directed edges $\alpha$, and define ${\bf A}_{\alpha} \equiv \{ A \mid (AB)_{\to} \in \alpha \}$.
As before, we assume if $(AW)_{\to} \in \alpha$, then for all $V \in \ch_{\cal G}(A)$, $(AV)_{\to} \in {\alpha}$.
Define
${\mathfrak f}_{\alpha} \equiv \{ f^{(A_iW)_{\to}}_{A_i} : {\mathfrak X}_{H_i} \mapsto {\mathfrak X}_{A_i} \mid (A_iW)_{\to} \in \alpha \}$
as the set of policies associated with edges in $\alpha$.  Note that ${\mathfrak f}_{\alpha}$ may contain multiple policies for a given treatment variable $A$.

Define $Y({\mathfrak f}_{\alpha})$, the counterfactual response of $Y$ to the set of path-specific policies $\mathfrak{f}_{\alpha}$, as the following generalization of (\ref{eqn:rec-sub-e}) and (\ref{eqn:rec-sub-f}):
{\small
\begin{align}
\label{eqn:rec-sub-ef}
Y(\{ f^{(A_iY)_{\to}}_{A_i}(H_i({\mathfrak f}_{\alpha})) | (A_iY)_{\to} \in \alpha \}, \{ \pa_{\cal G}^{\bar{\alpha}}(Y) \}({\mathfrak f}_{\alpha}) )
\end{align}
}
To reformulate our earlier example, if
$\tilde{f}^{(AM)_{\to}}_A$ assigns $A$ to a constant value $a'$, and
${\mathfrak f}_{\{ (AY)_{\to}, (AM)_{\to} \}} \equiv \{ f^{(AY)_{\to}}_A(W), \tilde{f}^{(AM)_{\to}}_A \}$, then $Y({\mathfrak f}_{\{ (AY)_{\to}, (AM)_{\to} \}}) \equiv Y(f^{(AY)_{\to}}_A(W), M(a',W), W)$.

The joint counterfactual distribution for responses to path-specific policies, $p(\{ V({\mathfrak f}_{\alpha}) | V \in {\bf V} \setminus {\bf A}_{\alpha} \})$, is identified under the functional model, and generalizes (\ref{eqn:g-e}) and (\ref{eqn:g-f}) as follows:
{\small
\begin{align}
\prod_{V \in {\bf V} \setminus {\bf A}_{\alpha}} p(V | \{ f^{(A_iV)_{\to}}_{A_i}(H_i) | (A_iV)_{\to} \in \alpha \}, \pa_{\cal G}^{\bar{\alpha}}(V))
\label{eqn:g-fe}
\end{align}
}
For example, $p(Y(f_A(W), M(a',W), W))$ is identified as 
$\sum_{W,M} p(Y | f_A(W), M, W) p(M | a', W) p(W)$ in Fig.~\ref{fig:triangle} (a).
We prove a general version of this result in the Appendix. A general identification theory for responses to path-specific policies can be found in \citep{Shpitser18IDpolicies}. 

\section{LEARNING OPTIMAL OVERALL POLICIES}
\label{sec:opt-overall-policy}

The goal of this paper is developing methods for learning optimal path-specific policies, in cases where responses to such policies are identified.  Before discussing optimal path-specific policies, we first review existing approaches to finding optimal overall policies.
Optimality may be quantified in a number of ways.  The set of optimal policies is commonly defined as
${\bf f}^*_{\bf A} \equiv \argmax_{{\bf f}_{\bf A}} \E[Y({\bf f}_{\bf A})]$. 

We will discuss existing methods for finding optimal policies ${\bf f}^*_{\bf A}$ in the context of a causal model implying \emph{positivity and sequential ignorability} \citep{robins86new}.  This model is graphically represented, for two time points, in Fig.~\ref{fig:triangle} (b), where $W_0$ are baseline factors, $W_1$ and $W_2$ are intermediate and final outcomes, $A_1,A_2$ are treatments.
Variables $W_0,W_1,W_2$ may be confounded by a hidden common cause $U$.  It is well known that in this model, $\E[Y({\bf f}_{\bf A})]$ is identified via (\ref{eqn:g-f}).  We now discuss a number of approaches for computing the optimal set ${\bf f}^*_{\bf A}$ given this identifying formula.


\subsubsection*{Approaches Based on Backwards Induction}
\label{sec:parametric}

In sequential decision problems, choosing optimal functions ${\bf f}_{\bf A}$ may appear to be a difficult search problem over a large set of function combinations.  However, it is possible to use ideas from dynamic programming to transform the problem of choosing optimal ${\bf f}_{\bf A}$ into a sequential problem where only a single optimal function is chosen at a time.

Multiple modeling approaches are possible here.  A simple approach is to model conditional densities of all outcomes $W_i$ given their past.  Assuming all such models were correctly specified, given any particular history $H_K$ up to the last decision point $A_K$, the optimal $f^*_{A_K}$ is equal to $\mathbb{I}(\E[ W_K(A_K=1) | H_K] > \E[ W_K(A_K=0) | H_K])$, which under our model is equal to
$\mathbb{I}(\E[ W_K | A_K=1, H_K] > \E[ W_K | A_K=0, H_K])$.
Assuming optimal $f^*_{A_{i+1}}, \ldots f^*_{A_K}$, denoted by $f^*_{\underline{A}_{i+1}}$, were already chosen, the optimal $f^*_{A_i}$ is defined inductively as
{\small
\begin{align*}
\mathbb{I}
\left[
\E[ W_K(A_i=1, f^*_{\underline{A}_{i+1}}) | H_{i}] > 
\E[ W_K(A_i=0, f^*_{\underline{A}_{i+1}}) | H_{i} ]
\right].
\end{align*}
}
%
\noindent Note that under our model, the counterfactual expectations above are identified via a modification of (\ref{eqn:g-f})
where the outer summation is only with respect to variables $W_{i+1}, \ldots, W_{K-1}$.
For many kinds of statistical models for densities appearing as terms in (\ref{eqn:g-f}),
evaluating this sum may be challenging.
An alternative strategy that avoids repeated summations is modeling the above expectations directly via \emph{Q-functions} $Q_i$, which are conditional expectations over \emph{value functions} $V_{i+1}$, given the history.  These are defined recursively as follows:
{\scriptsize
\begin{align*}
Q_K(H_K, A_K; \gamma_K) &= \E[W_K
 \mid A_K,
H_K
],  \\
V_K(H_K) &= \max_{a_K} Q_K(H_K, a_K;  \gamma_K),  
\end{align*}
}
and for $i = K-1, \ldots, 1$, as 
{\scriptsize
\begin{align*}
Q_i(H_i, A_i; \gamma_i) &= \E[V_{i+1}(H_{i+1}) \mid
A_i, H_i],   \\
V_i(H_i) &= \max_{a_i} Q_i(H_i, a_i; \gamma_i).  
\end{align*}
}
The optimal policy at each stage may be easily derived from Q-functions as
$f^*_{A_i}(H_i) = \argmax_{a_i} {Q}_i(H_i, a_i; \gamma_i)$.
Q-functions are recursively defined regression models where outcomes are value functions, and features are histories up to the current decision point.  Thus, parameters $\gamma_i, i=1, \ldots, K$, of all Q-functions may be learned recursively by maximum likelihood methods applied to regression at stage $i$, given that the value function at stage $i+1$ was already computed for every row \citep{Moodie13DTR, Schulte14DTR}.

\subsubsection*{Value Search}
\label{sec:value}

Consider a restricted class of policies $\cal F$ with elements
${\bf f}_{\bf A} \equiv \{ f_{A_i}(H_i); A_i \in {\bf A} \}$.
It is often of interest to estimate the optimal policy within the class ${\cal F}$, even if the class does not contain the true optimal policy.
For example, ${\cal F}$ may be the set of clinically interpretable policies.
For a sufficiently simple class ${\cal F}$, we can directly search for the optimal
${\bf f}^{*,{\cal F}}_{\bf A} \equiv \argmax_{{\bf f}_{\bf A} \in {\cal F}} \E[Y({\bf f}_{\bf A})]$.
This is called policy search or value search.

The expected response to an arbitrary treatment policy
$\beta = \E[Y({\bf f}_{\bf A})]$, for ${\bf f_A} \in {\cal F}$
can be estimated in a number of ways.  For ${\bf A} = \{ A \}$,
a simple estimator for $\beta$ that uses only the treatment assignment model $\pi(H; \psi)$ for $p(A=1 | H)$ is the inverse probability weighting (IPW) estimator:
{\small
\begin{align}
\label{eqn:ipw}
\E\left[ { Y C_{f_A}}/{ \pi_{f_A}(H; \widehat{\psi}) } \right],
\end{align}
}\noindent
where $\small C_{f_A} \equiv \mathbb{I}(A = f_A(H))$,
$\pi_{f_A}(H;\psi) \equiv \pi(H;\psi) f_A(H) + (1-\pi(H;\psi)) (1 - f_A(H))$,
the expectation is evaluated empirically, and $\widehat{\psi}$ is fit by maximum likelihood.
This estimator will not in general yield the optimal policy within ${\cal F}$ if $\pi$ is misspecified.  An alternative estimator for $\beta$ that provides some protection against this is the following: 
{\scriptsize
\begin{align}
& \E\left[ \frac{C_{f_A} Y}{\pi_{f_A}(H; \hat{\psi})} - \frac{C_{f_A} - \pi_{f_A}(H; \hat{\psi})}{\pi_{f_A}(H; \hat{\psi})}\E[Y | H, A=f_A(H);\hat{\gamma}] \right].
\label{eqn:robust}
\end{align}
}\noindent
The above estimator is \textit{doubly-robust} meaning that it is a consistent estimator if either the propensity score model $\pi(H; {\psi})$ or the regression model $\E[Y | H, A;{\gamma}]$ is correctly specified. 

Value search methods can in some cases be rephrased as a weighted classification problem in machine learning \citep{zhao12owl}.  In the interest of space, we do not discuss methods based on this observation further here.

\subsubsection*{G-Estimation}
\label{sec:g-estimation}

An alternative method for learning policies is to directly model the counterfactual contrast functions known as \emph{optimal blip-to-zero functions}, or the counterfactual deviations in outcome from a reference treatment value (which we take to be $A=0$), conditional on history, assuming all future decisions are already optimal.  Specifically, for each decision point $i$, we posit a \emph{structural nested mean
model (SNMM)} $\gamma_i(H_i, A_i; \psi)$ for the contrast
{\small
\begin{align*}
& \E\left[ Y(\bar{a}_i, f^*_{\underline{A}_{i+1}}) \mid H_i \right] - \E\left[ Y(\bar{a}_{i-1}, a_i = 0, f^*_{\underline{A}_{i+1}}) \mid H_i \right]. \nonumber 
\end{align*}
}
Note that if the true $\gamma_i(H_i,A_i;\psi)$ were known,
the optimal treatment policies are those that maximize the blip function at each stage:
$f^*_{A_i} = \argmax_{a_i} \gamma_i(H_i, A_i; \psi_i).$ 
In order to estimate $\psi$ using data, let 
{\scriptsize
\begin{align}
U({\bf \psi, \zeta(\psi), \alpha}) = \sum_{i=1}^{K} \left\{ G_i({\bf \psi}) - \E\left[G_i({\bf \psi}) \mid H_i; {\bf \zeta} \right] \right\} \ \nonumber \\
\times \left\{ d_i(H_i, A_i) - E\left[d_i(H_i, A_i) \mid H_i; {\bf \alpha}\right] \right\}, 
\label{eq:G-estim}
\end{align} 
}
where $d_i(H_i, A_i)$ is any function of $H_i$ and $A_i$, and 
{\scriptsize
\[
G_i({\bf \psi}) = Y  - \gamma_j(H_i, A_i; {\bf \psi}) +\!\!\! \sum_{k = i + 1}^{K} \left[ \gamma_k(H_k, a^*_k; {\bf \psi}) - \gamma_k(H_k, a_k; {\bf \psi}) \right].
\]}
Consistent and asymptotically normal (CAN) estimators of ${\bf \psi}$ can be obtained using the estimating equations $\E[U({\bf \psi, \zeta(\psi), \alpha})] = 0$, as shown in \citep{robins04SNMM}. The estimate obtained in (\ref{eq:G-estim}) is doubly-robust, meaning that the estimator $\widehat{\psi}$ is consistent if either $\E\left[G_i({\bf \psi}) \mid H_i; {\bf \zeta} \right]$ or $p_i(A_i = 1 \mid H_i; {\bf \alpha})$ is correctly specified.

Methods closely related to G-estimation based on \emph{counterfactual regret} were developed in \citep{murphy03optimal}.  
We do not discuss them here in the interest of space. 

\section{LEARNING OPTIMAL PATH-SPECIFIC POLICIES}
\label{sec:opt-edge-policy}

We now consider how the methods for optimizing overall policies translate to optimizing path-specific policies. Some of the generalizations we consider are currently only known for single-stage decision problems.

Consider the generalization of Fig.~\ref{fig:triangle} (b) to the longitudinal setting with mediators, shown (for two time points) in
Fig.~\ref{fig:swig} (a).  This causal model corresponds to the setting described in detail in \citep{miles17quantifying}, representing
an observational longitudinal study of HIV patients. Here, $W_0$ represents the baseline variables of a patient, $A_1,A_2$ represent
treatment assignments, which were chosen based on observed treatment history according to physician's best judgement, $W_1,W_2$
are intermediate and final outcomes (such as CD4 count or viral failure), and $M_1,M_2$ are measures of patient adherence to their
treatment regimen.  We are interested in finding policies $f_{A_1}(H_1), f_{A_2}(H_2)$ 
that optimize the effect of
$A_1, A_2$ on $W_2$ that is either direct or via intermediate outcomes, but not via adherence, and where adherence is kept to that of
a reference treatment $a'_1, a'_2$.  Specifically, we are interested in choosing
$f_{A_1},f_{A_2}$ to optimize the counterfactual expectation $\E[W_2(f_{A_1}, f_{A_2})]$, which expands via (\ref{eqn:rec-sub-ef}) to 
{\scriptsize
\begin{align}
\E\left[W_2\left(
	\begin{array}{c}
	W_0, 
	f_{A_1}(H_1), 
	M_1(a'_1), \\ 
	W_1\Big(f_{A_1}(H_1), M_1(a'_1)\Big), 
	f_{A_2}(H_2), \\  
	M_2\Big(a'_1, a'_2, W_1\big(f_{A_1}(H_1), M_1(a'_1)\big), M_1(a'_1)\Big) 
	\end{array}
	\right)\right].
	\label{eqn:2-stage}
\end{align}
}
The $K$ stage version of the causal model in Fig.~\ref{fig:swig} (a) is given by a complete DAG on variables
$W_0, A_1, M_1, W_1, \ldots, A_K, M_K, W_K$, listed in topological order, with a hidden common cause $U$ of $W_0, \ldots, W_K$.
Let $\alpha$ be all directed edges out of $A_1, \ldots, A_K$.
The general version of (\ref{eqn:2-stage}) is the expectation of $W_K$ taken with respect to the distribution
$p(W_K({\mathfrak f}_{\alpha}))$, where ${\mathfrak f}_{\alpha}$ sets all edges $(A_iM_j)_{\to}$ to $a'_i$, and all other edges in
$\alpha$ to a policy $f_{A_i}(H_i)$.

Identifiability of $p(W_K({\mathfrak f}_{\alpha}))$ is given by the following corollary of results in \citep{shpitser13cogsci}, which can be viewed as a generalization of the \emph{collapse of the g-formula}
\citep{robins86new} to longitudinal mediation settings.
\begin{thm}
In the above model, with a positive observed data distribution $p(W_K({\mathfrak f}_{\alpha}))$ is identified as
{\scriptsize
\begin{align}
\sum_{H_K,M_K}  \!\!\!\!\! 
\begin{array}{c} 
	\prod_{i=1}^K \Big\{  p(W_i | \overline{M}_{i-1}, \overline{W}_{i-1}, f_{\underline{A}_i}(H_i)) \\
		p(M_i | \bar{a}'_i, \overline{W}_{i-1}, \overline{M}_{i-1}) \Big\} \ p(W_0) 
\end{array}
\label{eqn:id-g-m}
\end{align}
}
\label{thm:collapse-g-e}
\end{thm}
As a consequence (\ref{eqn:2-stage}) is identified as
%
{\scriptsize
\begin{align*}
\sum_{\overline{W}_1, \overline{M}_2}
\!\!\!\!
\begin{array}{c}
\E\left[ W_2 |
f_{A_1}(H_1), f_{A_2}(H_2), \overline{W}_1,  \overline{M}_2  
\right] p(M_1 | a'_1,W_0) \times\\
\ p(W_1 | W_0, f_{A_1}(H_1), M_1) 
\ p(M_2 | \overline{W}_1, M_1, a'_1, a'_2) \ p(W_0).
\end{array}
\end{align*}
}
We now discuss a number of strategies for finding optimal path-specific policies identified by (\ref{eqn:id-g-m}).

\subsubsection*{Parametric Backwards Induction}

A simple approach for finding optimal path-specific policies is to combine backwards induction with a maximum likelihood estimator for the conditional densities in the identifying functional (\ref{eqn:id-g-m}). 


Consider Fig.~\ref{fig:swig} (a), and let 
$a_i$ and $a_i'$ denote the active and reference levels of $a_i$, respectively.  
Beginning at the last stage $K = 2$, and assuming treatment is binary, the optimal decision, $f^*_{A_2}$, for a given patient with history $(w_0, a_1, m_1, w_1)$ sets $A_2$ to either $a_2$ or $a_2'$ to maximize the response $W_2$, while keeping $M_2$ at whatever value it would have attained under a sequence of reference interventions
$(a_1',a_2')$:
{\scriptsize
\begin{align*}
f^*_{A_2} = \mathbb{I}\Big(\E[W_2(a_2, M_2(\overline{a}_2')) \mid H_2] >  \E[ W_2(a_2', M_2(\overline{a}_2')) \mid H_2] \Big).   
\end{align*}
}
The optimal decision at the first stage, $f^*_{A_1}$, is given by 
{\scriptsize
\begin{align*}
\mathbb{I}\Bigg\{ \E\left[W_2\left(
	\begin{array}{c}
	a_1,
	f^*_{A_2}(H_2), 
	M_1(a_1'),
	W_1(a_1, M_1(a_1')),	\\
	M_2(\overline{a}_2', M_1(a_1'), W_1(a_1, M_1(a_1')))
	\end{array}
	\right) \Big| H_1 \right] >  \nonumber \\
\E\left[W_2\left(
	\begin{array}{c}
	a_1',
	f^*_{A_2}(H_2), 
	M_1(a_1'),
	W_1(a_1', M_1(a_1')), \\
	M_2(\overline{a}_2', M_1(a_1'), W_1(a_1', M_1(a_1')))
	\end{array}
	\right) \Big| H_1 \right] \Bigg\}.
\end{align*}
}
Under the causal model we described, the above counterfactuals can be estimated using a modification of
(\ref{eqn:id-g-m}) with no summations over, but instead conditioning on histories $H_1$ or $H_2$.
This approach easily generalizes to any number of decision stages.  The difficulty here, as before, is the increasing amount of
marginalizations that must be performed as the number of stages grows.

\subsubsection*{Path-Specific Policies Via Q-Learning}

\begin{figure}
\begin{center}
\begin{tikzpicture}[>=stealth, node distance=1.0cm]
    \tikzstyle{format} = [draw, very thick, circle, minimum size=5.0mm,
	inner sep=0pt]
	\begin{scope}[xshift=0cm]
		\path[->, very thick]
			node[format] (w0) {$W_0$}
			node[format, below right of=w0] (a1) {$A_1$}
			node[format, above right of=w0] (m1) {$M_1$}
			node[format, below right of=m1] (w1) {$W_1$}
			node[format, gray, above right of=m1] (u) {$U$}			
			node[format, below right of=w1] (a2) {$A_2$}
			node[format, above right of=w1] (m2) {$M_2$}
			node[format, below right of=m2] (w2) {$W_2$}
			(w0) edge[blue] (a1)
			(w0) edge[blue] (m1)
			(w0) edge[blue] (w1)
			(a1) edge[blue] (m1)
			(a1) edge[blue] (w1)			
			(m1) edge[blue] (w1)
			(w1) edge[blue] (a2)
			(w1) edge[blue] (m2)
			(w1) edge[blue] (w2)
			(a2) edge[blue] (m2)
			(a2) edge[blue] (w2)			
			(m2) edge[blue] (w2)
			(u) edge[red, bend right=35] (w0)
			(u) edge[red, bend right=0] (w1)
			(u) edge[red, bend left=35] (w2)
			(m1) edge[blue] (m2)
			(a1) edge[blue] (a2)
			(a1) edge[blue, bend right=0] (w2)
			
			(a1) edge[blue, bend right=28] (m2)
			(m1) edge[blue, bend right=28] (a2)
			(m1) edge[blue] (w2)
			
			(w0) edge[blue] (m2)
			(w0) edge[blue, bend right=66] (a2)
			
			node[below of=a1, yshift=0.2cm, xshift=0.7cm] (l) {$(a)$}
		;
	\end{scope}
	\begin{scope}[xshift=4.0cm]
		\path[->, very thick]
			node[format] (w0) {$W_0$}
			node[format, below right of=w0] (a1) {$A_1$}
			node[format, above right of=w0] (m1) {$\tilde{M}_1$}
			node[format, below right of=m1] (w1) {$\tilde{W}_1$}
			node[format, gray, above right of=m1] (u) {$U$}			
			node[format, below right of=w1] (a2) {$\tilde{A}_2$}
			node[format, above right of=w1] (m2) {$\tilde{M}_2$}
			node[format, below right of=m2] (w2) {$\tilde{W}_2$}
			(w0) edge[blue] (a1)
			(w0) edge[blue] (m1)
			(w0) edge[blue] (w1)
			(a1) edge[blue] (w1)			
			(m1) edge[blue] (w1)
			(w1) edge[blue] (a2)
			(w1) edge[blue] (m2)
			(w1) edge[blue] (w2)
			(a2) edge[blue] (w2)			
			(m2) edge[blue] (w2)
			(u) edge[red, bend right=35] (w0)
			(u) edge[red, bend right=0] (w1)
			(u) edge[red, bend left=35] (w2)
			(m1) edge[blue] (m2)
			(a1) edge[blue] (a2)
			(a1) edge[blue, bend right=0] (w2)
			
			(m1) edge[blue, bend right=28] (a2)
			(m1) edge[blue] (w2)
			
			(w0) edge[blue] (m2)
			(w0) edge[blue, bend right=66] (a2)

			node[above left of=m1,yshift=-0.0cm] (a1') {$a'_1$}
			node[above right of=m2,yshift=-0.0cm] (a2') {$a'_2$}
			
			(a1') edge[blue] (m1)
			(a1') edge[blue] (m2)
			(a2') edge[blue] (m2)
			
			node[below of=a1, yshift=0.2cm, xshift=0.7cm] (l) {$(b)$}
		;
	\end{scope}
\end{tikzpicture}
\end{center}
\caption{
(a) A causal model that generalized Fig.~\ref{fig:triangle} (b) by also considering mediators.
(b) A ``multiple world intervention graph,'' representing the counterfactual situation where adherence levels are kept to a reference treatment level, but the chemical effect of drugs is operating normally.
}
\label{fig:swig}
\end{figure}
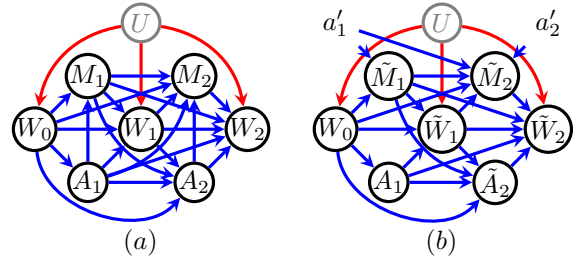

We now describe how to generalize Q-learning to path-specific policies, using the HIV example in Fig.~\ref{fig:swig} (a) to ground the discussion.  Recall that in this example, we wish to set $A_1$ and $A_2$ with respect to edges into $W_1,W_2$ to maximize the outcome, while setting $A_1$ and $A_2$ to reference values $a'_1,a'_2$ for the purposes of edges into $M_1, M_2$. 

Simply defining Q-functions as expectations over value functions conditional on history does not work in our setting, since mediators behave in a counterfactually different way from either observed variables, or variables we wish to set according to a policy. Moreover, the sequential nature of the problem means the nested counterfactuals needed become quite involved to write down. 

An alternative is to define Q-functions not in the observed data distribution, corresponding to Fig.~\ref{fig:swig} (a), but in a counterfactual distribution where $A_1,A_2$ behave as observed, \emph{except} for the purposes of edges into $M_1,M_2$, in which case they are counterfactually set to $a'_1,a'_2$.  The graph corresponding to this counterfactual world is shown in Fig.~\ref{fig:swig} (b), and can be viewed as a generalization of a single world intervention graph \citep{thomas13swig}, where treatment variables are only intervened on for the purposes of certain outgoing edges.  Note that descendant variables of $M_1,M_2$ are marked with a tilde to make clear that these variables are counterfactual and no longer equal to their observed counterparts.

The distribution corresponding to this situation is simply $p({\bf V}({\mathfrak a}_{(A_1M_1)_{\to},(A_1M_2)_{\to},(A_2M_2)_{\to}}))$, where
${\mathfrak a}$ sets these edges to $a'_1,a'_2$. 
For $K$ stages, the distribution is $p({\bf V}({\mathfrak a}_{\alpha}))$, where $\alpha$ are all edges of the form $(A_iM_j)_{\to}$, and
${\mathfrak a}$ sets each such edge to the reference value $a'_i$.  We have the following claim.
\begin{thm}
In $K$ stage version of the model in Fig.~\ref{fig:swig}(a), $p({\bf V}({\mathfrak a}_{\alpha}))$ is identified as
{\scriptsize
\begin{align}
\tilde{p}(\tilde{W}_0, \tilde{A}_1, \tilde{M}_1, \tilde{W}_1, \ldots, \tilde{W}_K, \tilde{A}_K, \tilde{M}_K) = \nonumber\\
p(W_0) \prod_{i=1}^K p(W_i | M_i, A_i, H_i) p(A_i | H_i) p(M_i | \overline{a}'_i, H_i \setminus {\bf A})
\label{eqn:edge-dist}
\end{align}
}
\label{thm:id-tilde}
\end{thm} 
\vspace{-0.3cm}
We can now define Q-functions as value function expectations on this new distribution, and proceed with backwards induction as before.  The only difference between the previous formulation is how Q-functions parameters are fit.  In particular, we must compensate for the fact that $\tilde{p}$ above is not the observed data distribution.  Define
{\scriptsize
\begin{align}
\tilde{Q}_K(\tilde{H}_K, \tilde{A}_K;\gamma_K) &= \tilde{\E}[\tilde{W}_K
 \mid 
\tilde{A}_K, \tilde{H}_K
], \nonumber \\
\tilde{V}_K(\tilde{H}_K) &= \max_{a_K} \tilde{Q}_K(\tilde{H}_K, {a}_K;  \gamma_K),  
\end{align}
}
and for $i = K-1, \ldots, 1$, define
{\scriptsize
\begin{align}
\tilde{Q}_i(H_i, A_i;\gamma_i) &= \tilde{\E}[\tilde{V}_{i+1}(\tilde{H}_{i+1}) \mid \tilde{A}_i, \tilde{H}_i],  \nonumber \\
\tilde{V}_i(\tilde{H}_i) &= \max_{a_i} \tilde{Q}_i(\tilde{H}_i, a_i;  \gamma_i).  
\end{align}
}
In our example in Fig.~\ref{fig:swig} (a), $K=2$, $H_1 \equiv \{ W_0 \}$,
$\tilde{H}_2 \equiv \{ W_0, M_1, W_1 \}$, and $\tilde{\E}$ denotes expectations with respect to appropriate
conditional distributions derived from $\tilde{p}$. 
Q-functions defined in this way can be used to obtain the optimal path-specific policy at each stage.
\begin{thm}
Given that each $\tilde{Q}_i, i=1,\ldots,K$ is specified correctly,
the optimal treatment at stage $i$ given $H_i$ is equal to:
$f^*_{A_i}(H_i) = \argmax_{a_i} \tilde{Q}_i(H_i, a_i; {{\gamma}_i})$.
\label{thm:f-star}
\end{thm}
\vspace{-0.4cm}
Since parameters $\gamma$ are not generally known, they must be estimated from data.
This can be done as follows.
 \begin{thm}
Assume models in the set $\{ \tilde{Q}_i(\tilde{H}_i, \tilde{A}_i; \gamma_i), p(M_i | A_i, H_i; \phi) |  \forall i \}$ are correctly
specified.
Then the estimation equations
{\scriptsize
\begin{align*}
\E\left[
\frac{\partial \tilde{Q}_K
}{\partial \gamma_K} \{ W_K - \tilde{Q}_K(A_K,H_K; \gamma_K)
\} w_K(H_K; \widehat{\phi_K})
\right] = 0, \text{  and }\\
\E\left[
\frac{\partial \tilde{Q}_i
}{\partial \gamma_i}
\{ V_{i+1}(H_{i+1}) -
\tilde{Q}_i(H_i, A_i; \gamma_i)
\}
w_i(H_i; \widehat{\phi_i})
\right] = 0,
\end{align*}
}
are consistent for $\gamma_K$ and $\gamma_i$, where
{\scriptsize
\begin{align*}
w_i(H_i; \widehat{\phi_i}) \equiv \frac{p(M_i | \overline{A_i}=a',H_i;\widehat{\phi}_i)}{p(M_i | \overline{A_i},H_i;\widehat{\phi_i})} \forall i = 1, \ldots K.
\end{align*}
}
\label{thm:q-learn-consist}
\end{thm} 

\subsubsection*{Path-Specific Value Search}

For simplicity, we restrict attention to a single-stage decision problem, as shown in Fig.~\ref{fig:triangle} (a), where we are interested in
picking a policy that maximizes the counterfactual mean $\beta = \E[Y(A=f(W), M(a'))]$.

Consider a restricted class of path-specific policies $\cal F$ with elements
${\mathfrak f} \equiv \{ f(W), \tilde{f} \}$ the latter setting $A$ to a constant value $a'$, regardless of $W$.
Give any estimation strategy for the counterfactual mean under a path-specific policy, we can
implement a direct search for the optimal policies within ${\cal F}$ as before.
By analogy with earlier discussion of value search,
we give an IPW estimator for $\beta$ which generalizes (\ref{eqn:ipw}):
{\scriptsize
\begin{align}
\E\left[
\frac{
Y \widetilde{C}
}{
\widetilde{\pi}(W; \widehat{\psi})
}
\cdot
\frac{
p(M | A=a', W; \widehat{\phi})
}{
p(M | A=f(W), W; \widehat{\phi})
}
\right],
\label{eqn:ipw-e}
\end{align}
}
and an estimator
which generalizes (\ref{eqn:robust}):
{\scriptsize
\begin{align}
& \E\left[
\frac{\widetilde{C}}{\widetilde{\pi}(W; \psi)} \frac{f(M | W, A = a'; \widehat{\phi})}{f(M | W, f(W); \widehat{\phi})} \Big\{ Y -
\mathbb{E}[Y | f(W),M,W; \widehat{\zeta}] \Big\} \ +  \right. \nonumber \\
&
\frac{\mathbb{I}(A=a')}{\pi_{a'}(W; \widehat{\psi})}
\Big\{ \mathbb{E}[Y | f(W),M,W; \widehat{\zeta}]
- \sum_{M} \mathbb{E}[Y | f(W),M,W; \widehat{\zeta}] \nonumber \\
& \left. p(M | W, A = a'; \widehat{\zeta})\Big\} + \sum_{M} \mathbb{E}[Y | f(W), M, W; \widehat{\zeta}] p(M | W, A = a'; \widehat{\phi})
\right],
\label{eq:triple-robust}
\end{align}
}\noindent
where the expectation is evaluated empirically,
$\widetilde{C} \equiv \mathbb{I}(A = f(W))$,
$\pi_{a'}(W;\psi) = p(A=a'|W;\psi)$,
$\widetilde{\pi}(W; \widehat{\psi}) \equiv \sum_{a} \pi_a(W; \psi) \mathbb{I}(a = f(W))$,
and $\widehat{\psi}$, $\widehat{\phi}$, $\widehat{\zeta}$ are fit by maximum likelihood.
We have the following. 
\begin{thm}
Under regularity assumptions referenced in the Appendix,
the estimator in (\ref{eqn:ipw-e}) is consistent and asymptocally normal (CAN) if the models in the set $\{ \pi(W; \psi), p(M | W, A; \phi) \}$ are correctly specified, and  the estimator in (\ref{eq:triple-robust}) is CAN in the \emph{union model}, where any two models in the set
$\{ \pi(W; \psi), \E[Y | A, M, W; \zeta], p(M | W, A; \phi) \}$ are correctly specified.
\label{thm:robust-1-a}
\end{thm}
This claim, and the estimators above, are extensions of the results in \citep{tchetgen12semi2}.

\subsubsection*{Single-Stage G-Estimation For Path-Specific Policies}

Results in \citep{tchetgen14semi} generalized optimal blip-to-zero functions to mediation settings, by positing the following SNMM
$\gamma(A, W; \psi)$:
{\scriptsize
\begin{align}
\E[ Y(A,M(A=0)) | W] - 
 \E[ Y(A=0,M(A=0)) | W].
\label{eq:Gestim-path}
\end{align}
}\noindent

Note that the policy $f^*_A$ maximizing $\E[Y(A=f^*_A(W), M(0))]$ can be directly obtained from $\gamma$ via
{\small
\[
f^*_A(W) = \argmax_{a} \gamma(A, W; \psi).
\]
}
Since $\psi$ is not generally known, it must be estimated from data.
The following is a consistent set of estimating equations for $\psi$, under assumptions described in \citep{tchetgen14semi}:
{\scriptsize
\begin{align}
\label{eqn:snmm-med}
\E\left[
\left(
\frac{\mathbb{I}(A=1) p(M | A=0,W)}{p(A=1|W) p(M | A=1,W)} \{ Y - \E[Y | A=1,M,W] \} - \right.\right. \\
\left.
\left.
\frac{
\mathbb{I}(A=0)
}{
p(A=0 | W)
} \{ Y - \E[ Y | A=1, M, W] + \gamma(1, W; \psi) \}
\right)
h(W)
\right] = 0 \nonumber
\end{align}
}
for $h(W)$ any $|\psi |$-dimensional function of $W$.

Note that unlike SNMMs associated with overall policies, which can be defined for any number of treatments, SNMMs associated with path-specific policies have only been defined for a single treatment. A longitudinal generalization of these models is {left to future work.} 

\section{EXPERIMENTS}
\label{sec:experiment}
We now illustrate our methods via a dataset on HIV patients from Nigeria. The data consists of more than $50k$ treatment-naive HIV-1 infected patients who were enrolled in the Harvard PEPFAR/AIDS Prevention Initiative program prior to Oct 2010. The patients were put on courses of antiretroviral therapy (ART), with five standard first-line regimen, and were followed every six months for at least one year after the ART initiation. Patients stayed on their initial treatment plan unless they were experiencing toxicity within a first-line drug regimen and consequently moved to a second-line regimen. The data has records on demographics and clinical test results such as CD4 counts, viral loads, and toxicity measures at 6-month intervals. 

In order to combat chronic diseases such as HIV, patients are required to follow long term use of medications. Full benefits of the medications are realized when patients take their medications as prescribed. Unfortunately, factors such as side effects, caused by drug toxicities, can be developed alongside the course of treatment and lead to poor adherence to the therapy. A primary measure of adherence is provided in the data as average percent adherence that is the total number of days that the patient has drug supply over the total number of days in the time period (6-month intervals). 

We restricted our attention to the first year of follow up, including two decision points and picked (log) CD4 count at the end of the first year as the outcome of interest. CD4 count is one of the biomarkers that quantifies how well the immune system is functioning and higher values are more favorable.
Drug toxicity and adherence can mediate the effect of treatment on outcome. Since reactions to drug toxicity and adherence behavior vary among patients, we consider the problem of finding policies that optimize the direct chemical effect of the drug where the indirect effect mediated through drug toxicity and adherence are set to some reference levels. {We run an additional experiment where only the effect through adherence is set to a reference level \cite{miles17quantifying}. The latter results are provided in the Appendix for the interest of space}. 

The causal model 
can be represented by the DAG in Fig.~\ref{fig:swig} (a). Treatment decisions at the baseline and the first follow up 
are respectively denoted by $A_1$ and $A_2$. 
$W_1$ is a dichotomized intermediate outcome that indicates whether the CD4 count is above $450$ cells/mm$^3$ at the the end of the same six month period. $W_2$ is the final outcome and denotes the log CD4 count at the end of the first year after ART initiation. $W_0$ includes all the baseline factors such as sex, age, and test results prior to any treatment initiation.  Toxicity and adherence measures during the first six months and the first year after treatment initiation are collected in $\bf M_1$ and $\bf M_2$, respectively. Drug toxicity indicates any lab toxicities (alanine transaminase $\geq 120$ UI/L, creatinine $ \geq 260$ mmol/L, hemoglobin $\leq 8$ g/dL), and adherence indicates whether average percent adherence was no less than $95\%$. 


We first illustrate the results on finding optimal policies in a two-stage decision problem using G-formula and Q-learning methods, and then provide the results for finding optimal policies for the single 
{(first)} stage decision problem using value search and G-estimation. {In the single stage analysis, we consider the log CD4 count at the end of the sixth month as the outcome of interest. All modeling assumptions are described in the Appendix.}

\subsubsection*{Methods For The Two-Stage Decision Problem}
%
Optimal overall polices and path policies are estimated as described in Sections~\ref{sec:opt-overall-policy} and~\ref{sec:opt-edge-policy}.
Expected outcomes under optimal policies we learned, along with 95\% confidence intervals obtained by bootstrap, are shown in Table.~\ref{tab:CD4}.  In the interest of space, we do not consider corrections necessary to preserve the validity of these intervals in cases of model discontinuities, but these are straightforward \cite{bibhas13inference}.
Both 
optimal polices have higher expected outcomes than the observed data, using either method.  Path-specific policies led to higher expected outcomes compared to overall policies.
This could be explained by the phenomenon of countervailing adherence-mediated effect, described in \citep{miles17quantifying}.
In other words, in some cases the more chemically effective drugs are also harder to take.
%
In order to visualize optimal policies we learned as clinically interpretable flowcharts, we viewed policy-recommended decisions as class labels, and history as features, and used a multilabel decision tree classifier, as implemented in the \texttt{rpart} R package.
The results are shown in Fig.~\ref{fig:Gformula} and \ref{fig:Qlearning} in the Appendix.  Since the classifiers are not perfect predictors of the policies, they are to be viewed as interpretable approximations of the true learned policy.  Variables involved in the decisions, such as age, gender, CD4, virological status, and so on, are clinically reasonable.

{\small
\begin{table}[t] 
\centering
\caption{{\small Comparison of population log CD4 counts under different policies (under treatment assignments in the observed data, the value is $5.64 \pm 0.01$ in the 2-stage and $5.54 \pm 0.01$ in the 1-stage problem). G-formula and Q-learning are used with 2-stage decision points. Value search and G-estimation are used with 1-stage decision point.}}
\scalebox{0.9}{
\begin{tabular}{c|c|c}
											  &          \small \textbf{Path Policies}           &         \small \textbf{Overall Policies}   \\ 
\hline 
\small \textbf{G-formula}      &  $ 6.89 \hspace{0.15cm} (5.76, 7.10) $ & $ 6.79 \hspace{0.15cm} (5.68, 6.90) $  \\
\small \textbf{Q-learning}     &   $ 7.34 \hspace{0.15cm} (6.10, 7.55) $ & $  6.89 \hspace{0.15cm} (5.82, 7.12) $ \\ 
\hline
\small \textbf{Value search}  &  $ 5.58 \hspace{0.15cm} (5.54, 5.62) $ & $ 5.56 \hspace{0.15cm} (5.55, 5.58) $ \\
\small \textbf{G-estimation}  &  $ 5.62 \hspace{0.15cm} (5.50, 5.67) $ & $ 5.79 \hspace{0.15cm} (5.59, 6.04) $
\end{tabular}
}
\label{tab:CD4}
\end{table}
}

\subsubsection*{Methods For The Single-Stage Decision Problem}
%
{We focus on patients that appear at baseline, and treat the five first-line treatments as a binary decision by grouping the first three (which we denote as group 1) and the last two treatments (which we denote as group 2).  We are interested in finding a treatment group assignment at the first decision point that leads to a higher CD4 count at the first follow up, assuming adherence and  toxicity are set to a reference point for every patient.}
The model for this setting is Fig.~\ref{fig:triangle}(a),
where $W$ is observed history up to the 
{first} decision.  

We considered policies of the form $\mathbb{I}\{\text{CD4m00} < \alpha\}$, where \textit{CD4m00} denotes the CD4 count right before the first decision point, to decide {what treatment group the patient should be assigned to}. 
The normal range for this variable is $500$ to $1500$ cells/mm$^3$.  We searched for an optimal policy in this restricted class by
varying $\alpha$ from $100$ cells/mm$^3$ to $1000$ cells/mm$^3$ by $50$ cells/mm$^3$ increments, and estimated the value for each $\alpha$ using (\ref{eq:triple-robust}).
Under the modeling assumptions described in the Appendix, the optimal threshold between group 1 and group 2 treatments was chosen to be $\mathbb{I}\{\text{CD4m00} < 550 \text{ cells/mm}^3 \}$. In other words, if the CD4 count is less than $550$, it is better for the patients to 
{receive any of the treatments in the first treatment group}. 
However, if we pick the optimal policy based on the overall effect of treatment on outcome, then value search leads to policies of the form $\mathbb{I}\{\text{CD4m00} < 250 \text{ cells/mm}^3 \}$.  The optimal overall policy decides to switch to group 2 treatments when HIV gets more severe, while the optimal path policy decides on switching when CD4 count is still within a healthy range but at lower values.  This implies that if treatment adherence could be kept to that of a reference treatment, initiation of treatments within group 2 could be delayed to lower CD4 values.  As before, both policies lead to a higher than observed log CD4 count, with the path policy yielding slightly higher outcomes than the overall policy.

Finally, we learned optimal path policies via G-estimation.
We estimated the parameters $\psi$ for the SNMM using (\ref{eqn:snmm-med}). The population log CD4 count under path policies and overall policies learned by G-estimation are given in Table.~(\ref{tab:CD4}), and a decision tree view of the policies is shown in Fig.~\ref{fig:g-estimation} in the Appendix.  Under our assumptions, the optimal path policy found by G-estimation did not do much better than the outcomes under observed treatments and the overall policies. 

\section{CONCLUSIONS}
\label{sec:conclusion}

In this paper, we generalized ideas in mediation analysis and dynamic treatment regimes to consider the problem of estimation of responses to policies associated with particular causal pathways, and methods for learning policies that optimize these responses.
Since validating findings in causal inference is difficult, and conclusions are sensitive to specific modeling assumptions made, we developed multiple approaches for learning optimal policies that rely on orthogonal sets of modeling strategies.  In particular,
we considered strategies based on backwards induction with either plug-in estimation or Q-learning, value search for restricted classes of policies, and G-estimation of structural nested mean models (SNMMs) generalized to mediation. 

We illustrated these methods by finding policies for HIV-positive patients that optimize the direct chemical effect of antiretroviral therapy, where the indirect effect mediated through drug toxicity and adherence are set to a reference level.  The results in the experiment section suggest that policies that aim to optimize the direct effect of the treatment have better outcome responses than policies that optimize the overall effect of the treatment, and optimal policies decide on clinically relevant variables, such as age, gender, viral load, and CD4 count.

The estimation methods we described are applicable to sequential decision problems, except for G-estimation which is currently only limited to single-stage decision problems.  Generalizing the version of G-estimation to longitudinal mediation problems, and deriving semi-parametric estimators for the version of Q-learning we described are areas for future work.

\subsection*{Acknowledgments}
This research was supported in part by the NIH grants R01 AI104459-01A1 and R01 AI127271-01A1. We thank the anonymous reviewers for their insightful comments that greatly improved this manuscript.

\bibliographystyle{abbrv} 
\bibliography{references}

\newpage
\section*{Appendix}

In Appendix A, we give a brief overview of semi-parametric statistical inference, which provides context for some of our subsequent results.  Appendix B contains deferred proofs of our results.  In Appendix C, we review the statistical modeling assumptions we made in our data analysis, provide figures which use decision tree classifiers to visualize policies we learned, and describe additional experimental results on policies that optimize effects not mediated by only adherence.

\subsection*{A: Statistical Inference In Semi-Parametric Models}
\label{subsec:statistical_inference}


Let {\small$Z_1, \ldots, Z_n$}, be iid samples from a general class of probability densities {\small$p(Z; \theta)$} parameterized by {\small$\theta^T = (\beta^T, \eta^T)$}, where {\small$\beta \in \mathbb{R}^q$} denotes the set of target parameters, and {\small$\eta$} denotes a possibly infinite dimensional set of nuisance parameters.  This type of model is termed semi-parametric, since it has both a parametric and a non-parametric component.  The goal of statistical inference in semi-parametric models is to find ``the best" estimator of {\small$\beta$} in the model, denoted by {\small$\widehat{\beta}$}.
\emph{Regular asymptotically linear (RAL)} estimators are considered in this setting, which are estimators of the form
\begin{eqnarray}
\sqrt{n}(\hat{\beta} - \beta) = \frac{1}{\sqrt{n}} \sum_{i = 1}^{n} \phi(Z_i) + o_p(1), \nonumber 
\end{eqnarray} 
where {\small$\phi \in \mathbb{R}^q$} with mean zero and finite variance, {\small$o_p(1)$} denotes a term that approaches to zero in probability, and
{\small$\phi(Z_i)$} is the \emph{influence function (IF)} of the {\small$i$}th observation for the parameter vector {\small$\beta$}.
RAL estimators are consistent and asymptotically normal (CAN), with the variance of the estimator given by its IF:
\begin{eqnarray}
\sqrt{n}(\hat{\beta} - \beta) \xrightarrow[]{\mathcal{D}} \mathcal{N}(0, \phi\phi^T). \nonumber 
\end{eqnarray}

Thus, there is a bijective correspondence between RAL estimators and IFs.  In fact, IFs provide a geometric view of the behavior of RAL estimators.
Consider a Hilbert space {\small${\cal H}$} of all mean-zero {\small$q-$}dimensional functions, equipped with an inner product, and define the inner product of two arbitrary elements of the Hilbert space, {\small$h_1$} and {\small$h_2$}, to be equal to {\small$\mathbb{E}[h_1^Th_2]$}.
Define a \emph{parametric submodel} to be a subset of densities in the semi-parametric model parameterized by {\small$\theta^T_{\gamma} = (\beta^T, \gamma^T)$}, where {\small$\gamma^T \in {\mathbb R}^r$}, such that the subset contains the density {\small$p(Z; \theta_0)$ in the semi-parametric model evaluated at the true parameter values {\small$\theta_0$}.
The \emph{nuisance tangent space} {\small$\Lambda$} in the semi-parametric model is defined to be the mean square closure of
elements of the nuisance tangent spaces {\small$\Lambda_{\gamma} = \{B^{q\times r} S_\eta(Z; \theta) \}$} of every parametric submodel.
The space {\small$\Lambda$} is important because it is known all influence functions lie in the orthogonal complement {\small$\Lambda^{\perp}$} of {\small$\Lambda$} with respect to {\small${\cal H}$}. For this reason, recovering {\small$\Lambda^{\perp}$} is often the first step for constructing RAL estimators in semi-parametric models.  Out of all IFs in {\small$\Lambda^{\perp}$} there exists a unique one which lies in the tangent space, and which yields the most efficient RAL estimator by recovering the \emph{semi-parametric efficiency bound}, see \citep{tsiatis06missing} for details.

\subsection*{B: Proofs}

Here we give proofs of all claims in the main body of the paper.

\begin{thma}{\ref{thm:collapse-g-e}}
Fix a causal model given by a complete DAG on variables
{\small$W_0, A_1, M_1, W_1, \ldots, A_K, M_K, W_K$}, listed in topological order, with a hidden common cause {\small$U$} of {\small$W_0, \ldots, W_K$}.
Let {\small$\alpha$} be all directed edges out of {\small$A_1, \ldots, A_K$}, and
{\small${\mathfrak f}_{\alpha}$} which sets all edges {\small$(A_iM_j)_{\to}$} to {\small$a_i$}, and all other edges in
{\small$\alpha$} to a policy {\small$f_{A_i}(H_i)$}.
In this model, {\small$p(W_K({\mathfrak f}_{\alpha}))$} is identified as
{\scriptsize
\begin{align}
\sum_{H_K,M_K}  \!\!\!\!\! p(W_0)
\prod_{i=1}^K & p(M_i | \bar{a}'_i, \overline{W}_{i-1}, \overline{M}_{i-1}) 
		 p(W_i | \overline{M}_{i-1}, \overline{W}_{i-1}, f_{\underline{A}_i}(H_i)) 
\label{eqn:id-g-m}
\end{align}
}
\end{thma}
\begin{prf}
The causal model we describe is simply Pearl's functional model corresponding to the K stage version of the DAG in Fig.~\ref{fig:swig} (a).
It is well known that in this model, given standard positivity assumptions, {\small$p(W_0, \ldots, W_K, M_1, \ldots, M_k | \text{do}(a_1, \ldots, a_K))$} is identified by the g-formula
(\ref{eqn:g}):
{\scriptsize
\begin{align*}
& p(W_0) \prod_{i=1}^K  p(M_i | \bar{a}'_i, \overline{W}_{i-1}, \overline{M}_{i-1}) p(W_i | \overline{M}_{i-1}, \overline{W}_{i-1}, \bar{a}_i).
\end{align*}
}
Since the recanting district criterion \citep{shpitser13cogsci} does not hold, we have
that {\small$p(\{ W_0, \ldots, W_K, M_1, \ldots, M_k \}({\mathfrak a}'_{\alpha}))$} is identified by
{\scriptsize
\begin{align*}
& p(W_0) \prod_{i=1}^K  p(M_i | \bar{a}'_i, \overline{W}_{i-1}, \overline{M}_{i-1}) p(W_i | \overline{M}_{i-1}, \overline{W}_{i-1}, \bar{a}_i).
\end{align*}
}
where {\small$\alpha$} is all outgoing edges from {\small${\bf A}$}, and {\small${\mathfrak a}'$} sets all edges of the form {\small$(A_iM_j)$} to {\small$a'_i$}, and
all edges of the form {\small$(A_iW_j)$} to {\small$a_i$}.

Every {\small$f_{A_i}(H_i)$} simply chooses {\small$a_i$} based on {\small$H_i$}, which is a subset of outcome variables in our distribution.
Since the identifiability statement above holds regardless of how {\small$a_1, \ldots, a_i$} are chosen, this
implies {\small$p(\{ W_0, \ldots, W_K, M_1, \ldots, M_k \}({\mathfrak f}_{\alpha}))$}, where {\small${\mathfrak f}_{\alpha}$} is given in the Theorem statement is identified as
{\scriptsize
\begin{align}
\sum_{H_K,M_K}  \!\!\!\!\! p(W_0)
\prod_{i=1}^K & p(M_i | \bar{a}'_i, \overline{W}_{i-1}, \overline{M}_{i-1}) 
		 p(W_i | \overline{M}_{i-1}, \overline{W}_{i-1}, f_{\underline{A}_i}(H_i)) 
\label{eqn:id-g-m}
\end{align}
}
which implies our result by a simple marginalization.
\end{prf}

Before proving Theorem \ref{thm:robust-1-a}, we show the following claim. 

\begin{thm}
Within the model corresponding to Fig.~\ref{fig:triangle} (a),
the unique \emph{efficient influence function} {\small$U(\beta)$} of {\small$\beta = \E[Y(A=f(W), M(a'))]$} is given
by 
{\scriptsize
\begin{align}
\label{eq:triple-robust}
& 
\frac{\widetilde{C}}{\widetilde{\pi}(W)} \frac{f(M | W, A = a')}{f(M | W, f(W))} \Big\{ Y -
\mathbb{E}[Y | f(W),M,W] \Big\} + \nonumber  \\
&
\frac{\mathbb{I}(A=a')}{\pi_{a'}(W)}
\Big\{ \mathbb{E}[Y | f(W),M,W]
- \sum_{M} \mathbb{E}[Y | f(W),M,W] \nonumber \\
&  p(M | W, A = a')\Big\} + \sum_{M} \mathbb{E}[Y | f(W), M, W] p(M | W, A = a') - \beta.
\nonumber
\end{align}
}
\end{thm}
\begin{prf}
This proof follows as an extension of similar results on the influence function of the mediation functional, found in \citep{tchetgen12semi2}.

The model in Fig.~\ref{fig:triangle} (a) imposes no restrictions on the observed data, and so is non-parametric saturated.  As a result, the influence function {\small$U(\beta)$} for any {\small$\beta$} is a unique solution to the following integral equation
{\small
\[
\left. \frac{\partial}{\partial t} \beta(F_t) \right|_{t=0} = \E[ S(W,A,M,Y) \phi(\beta) ],
\]
}
where {\small$F_t$} is the distribution function corresponding to a one dimensional regular parametric submodel of the non-parametric model on {\small$W,A,M,Y$}, indexed by a single parameter {\small$t$}, and {\small$S$} is the score.
{\small${\partial \beta(F_t)}/{\partial t} $} is equal to
{\scriptsize
\begin{align*}
& \frac{\partial}{\partial t} \sum_{w,m} \mathbb{E}[Y \mid a=f_A(w),m,w] p(m \mid a', w) p(w) =\\
& \sum_{w,m,y} y \frac{\partial}{\partial t} (p(y \mid a=f_A(w),m,w) p(m \mid a', w) p(w)) =\\
&
\sum_{w,m,y} y S(y \mid a=f_A(w),m,w) p(y \mid a=f_A(w),m,w)\times\\
& p(m \mid a', w) p(w) \\
&+ \sum_{w,m} \mathbb{E}[Y \mid a=f_A(w),m,w] S(m \mid a',w) p(m \mid a',w) p(w) \\
&+ \sum_{w,m}  \mathbb{E}[Y \mid a=f_A(w),m,w] p(m \mid a', w) S(w) p(w),
\end{align*}
}
where {\small$S(.)$} represent appropriate conditional and marginal scores.
By linearity of derivatives, we can solve this equation, term by term.
We have, for the first term:
{\scriptsize
\begin{align*}
& \sum_{w,m,y} y S(y \mid a=f_A(w),m,w) p(y \mid a=f_A(w),m,w)\times\\
& p(m \mid a', w) p(w)\\
&=
\sum_{w,m,y,a''} \frac{\mathbb{I}(a''=f_A(w)) p(m \mid a',w)}{p(a''=f_A(w) \mid w) p(m \mid a'', w)} \\
& y S(y \mid a'',m,w) p(y \mid a'',m,w) p(m \mid a'', w) \times\\
& p(a'' =f_A(w) \mid w) p(w)\\
&=
\sum_{w,m,y,a''} \frac{\mathbb{I}(a''=f_A(w)) p(m \mid a',w)}{p(a'' \mid w) p(m \mid a'', w)}\\
& \{ y - \mathbb{E}[Y \mid a'', m,w] \} S(y,a'',m,w) p(y,a'',m,w) \\
&=
\mathbb{E} \Big[
\frac{\mathbb{I}(A = f_A(W)) p(M \mid a', W)}{p(A=f_A(W) \mid W) p(M \mid A,W)} \times\\
& \hspace{1cm} \{ Y - \mathbb{E}[Y \mid A,M,W] \}  S(Y,A,M,W)
\Big].
\end{align*}
}
So the first term contribution to {\small$U(\beta)$} is {\small$\frac{\mathbb{I}(A = f_A(W)) p(M \mid a', W)}{p(A=f_A(W) \mid W) p(M \mid a,W)} \{ Y - \mathbb{E}[Y \mid a,M,W] \}$}.

For the second term, we have:
{\scriptsize
\begin{align*}
 \sum_{w,m} & \mathbb{E}[Y \mid a=f_A(w),m,w] S(m \mid a',w) \times\\
 &p(m \mid a',w) p(w)\\
=
\sum_{w,m,a''} & \frac{\mathbb{I}(a'' = a')}{p(a'' \mid w)} \mathbb{E}[Y \mid a=f_A(w),m,w]\times\\
 & S(m \mid a'',w) p(m,a'',w) \\
=
\sum_{w,m,a''} & \frac{\mathbb{I}(a'' = a')}{p(a'' \mid w)} \Big\{ \mathbb{E}[Y \mid a=f_A(w),m,w] \\
 & - \sum_{m} \mathbb{E}[Y \mid a=f_A(w),m,w] p(m \mid a'',w) \Big\} \times\\
 & S(m, a'',w) p(m,a'',w)\\
=
\sum_{w,m,a'',y} & \frac{\mathbb{I}(a'' = a')}{p(a'' \mid w)} \Big\{ \mathbb{E}[Y \mid a=f_A(w),m,w] \\
 & - \sum_{m} \mathbb{E}[Y \mid a=f_A(w),m,w] p(m \mid a'',w) \Big\} \times\\
 & S(y,m, a'',w) p(y,m,a'',w)\\
=
\mathbb{E}\Big[ &
S(Y,M,A,W)\frac{ \mathbb{I}(A = a')}{p(a' \mid W)} \times\\
& \left\{ \mathbb{E}[Y | a=f_A(W),M,W] - \mathbb{E}_q[Y | a,a',W] \right\} 
\Big].
\end{align*}
}
where {\small$\mathbb{E}_q[Y \mid a=f_A(W),a',W] = \sum_{m} \mathbb{E}[Y \mid a=f_A(W),m,W] p(m \mid a', W)$}.

So the second term contribution to {\small$U(\beta)$} is
{\scriptsize
\begin{align*}
\frac{\mathbb{I}(A = a')}{p(a' \mid W)} \Big\{ \mathbb{E}[Y \mid a=f_A(W),M,W] \\
- \mathbb{E}_q[Y \mid a=f_A(W),a',W] \Big\}.
\end{align*}
}

For the third term, we have:
{\scriptsize
\begin{align*}
& \sum_{w,m} \mathbb{E}[Y \mid a,m,w] p(m \mid a', w) S(w) p(w)\\
&=
\sum_{w} \left\{ \sum_{m} \mathbb{E}[Y \mid a,m,w] p(m \mid a', w) \right\} S(w) p(w)\\
&=
\sum_{w} \left\{ \sum_{m} \mathbb{E}[Y \mid a,m,w] p(m \mid a', w) \right.\\
& \left. - \sum_{w,m} \mathbb{E}[Y \mid a,m,w] p(m \mid a', w) p(w) \right\} S(w) p(w)\\
&=
\sum_{w,a'',m,y} \left\{ \sum_{m} \mathbb{E}[Y \mid a,m,w] p(m \mid a', w)\right. \\
& \left. - \sum_{w,m} \mathbb{E}[Y \mid a,m,w] p(m \mid a', w) p(w) \right\} \times\\
& S(y,m,a'',w) p(y,m,a'',w)\\
&=
\mathbb{E}\left[
\sum_{m} \mathbb{E}[Y \mid a=f_A(W),m,W] p(m \mid a', W) - \beta
\right]
\end{align*}
}
So the third term contribution to {\small$U(\beta)$} is {\small$\mathbb{E}_q[Y \mid a=f_A(W),a',W]  - \beta$}.

This establishes our result.
\end{prf}

\begin{thma}{\ref{thm:id-tilde}}
Fix a causal model given by a complete DAG on variables
{\small${\bf V} \equiv \{ W_0, A_1, M_1, W_1, \ldots, A_K, M_K, W_K \}$}, listed in topological order, with a hidden common cause {\small$U$} of {\small$W_0, \ldots, W_K$}.  Let {\small$\alpha$} be all directed edges present in the DAG out of {\small$A_1, \ldots, A_K$} and into {\small$M_1, \ldots M_K$}, and
{\small${\mathfrak a}_{\alpha}$} which sets all edges {\small$(A_iM_j)_{\to}$} to {\small$a'_i$}.
In this model, {\small$p({\bf V}({\mathfrak a}_{\alpha}))$} is identified as
{\scriptsize
\begin{align*}
p({\bf V}({\mathfrak a}_{\alpha})) \equiv \tilde{p}(\tilde{W}_0, \tilde{A}_1, \tilde{M}_1, \tilde{W}_1, \ldots, \tilde{W}_K, \tilde{A}_K, \tilde{M}_K) = \nonumber\\
p(W_0) \prod_{i=1}^K p(W_i | M_i, A_i, H_i) p(A_i | H_i) p(M_i | \overline{a}'_i, H_i \setminus {\bf A})
\end{align*}
}
\end{thma}
\begin{prf}
This is a corollary of Theorem \ref{thm:collapse-g-e}, where we define each {\small$f_{A_i}(H_i)$} to be the observed conditional distribution
{\small$p(A_i | H_i)$}.
\end{prf}

\begin{thma}{\ref{thm:f-star}}
Given that each {\small$\tilde{Q}_i, i=1,\ldots,K$} is specified correctly,
the optimal treatment at stage {\small$i$} is equal to
{\small
\begin{align*}
f^*_{A_i}(H_i) = \argmax_{a_i} \tilde{Q}_i(H_i, a_i; {{\gamma}_i}).
\end{align*}
}
\end{thma}
\begin{prf}
This follows by the standard backwards induction argument giving the relationship between Q-functions and optimal policies, applied to
{\small$\tilde{p}$} and {\small$\tilde{Q}_i$}, and the definition of expected response corresponding to path-specific policies we have chosen.

The optimal policy set {\small${\bf f}^*_{\bf A}$} is defined as
{\scriptsize
\begin{align*}
\argmax_{\{ f^*_{A_i} \in {\bf f}^*_{\bf A} \}} & \E[W_K({\mathfrak f}_{\alpha})]\\
= \argmax_{\{ f^*_{A_i} \in {\bf f}^*_{\bf A} \}} & \int W_K \prod_{i=1}^K p(W_i | \overline{A}_i = f^*_{\overline{A}_i}(H_i), H_i, M_i)\\
&p(W_0) p(M_i | \overline{a}'_i, H_i) d {\bf V}\\
= 
\argmax_{a_{1}} & \int p(W_{1} | a_{1}, H_{1}, M_{1}) 
 p(M_{1} | {a}'_{1}, H_{1}) dM_{1},W_{1} \\
& \ldots \\
\argmax_{a_{K-1}} & \int p(W_{K-1} | a_{K-1}, H_{K-1}, M_{K-1}) \\
& p(M_{K-1} | \overline{a}'_{K-1}, H_{K-1}) dM_{K-1},W_{K-1} \\
\argmax_{a_K} & \int W_K p(W_K | a_K, H_K, M_K)\\
& p(M_K | \overline{a}'_K, H_K) dM_K,W_K \\
\end{align*}
}

It's immediately clear that the last line above yields {\small$\tilde{Q}_K$}, and given that line {\small$i+1$} yields {\small$\tilde{Q}_{i+1}$} assuming {\small$a_{i+1}, \ldots, a_K$} were chosen optimally, line {\small$i$} yields {\small$\tilde{Q}_i$}.
\end{prf}

 \begin{thma}{\ref{thm:q-learn-consist}}
Assume models in the set $\{ \tilde{Q}_i(\tilde{H}_i, \tilde{A}_i; \gamma_i), p(M_i | A_i, H_i; \phi) |  \forall i \}$ are correctly
specified.
Then the estimation equations
{\scriptsize
\begin{align*}
\E\left[
\frac{\partial \tilde{Q}_K
}{\partial \gamma_K} \{ W_K - \tilde{Q}_K(A_K,H_K; \gamma_K)
\} w_K(H_K; \widehat{\phi_K})
\right] = 0, \text{  and }\\
\E\left[
\frac{\partial \tilde{Q}_i
}{\partial \gamma_i}
\{ V_{i+1}(H_{i+1}) -
\tilde{Q}_i(H_i, A_i; \gamma_i)
\}
w_i(H_i; \widehat{\phi_i})
\right] = 0,
\end{align*}
}
are consistent for $\gamma_K$ and $\gamma_i$, where
{\scriptsize
\begin{align*}
w_i(H_i; \widehat{\phi_i}) \equiv \frac{p(M_i | \overline{A_i}=a',H_i;\widehat{\phi}_i)}{p(M_i | \overline{A_i},H_i;\widehat{\phi_i})} \forall i = 1, \ldots K.
\end{align*}
}
\end{thma}

\begin{prf}
We show this inductively on the decision stage.  For stage {\small$K$}, we have
{\scriptsize
\begin{align*}
\E\left[ 
 \frac{\partial \tilde{Q}_K
 ; \gamma_K)}{\partial \gamma_K}
\{ W_K - \tilde{Q}_K(A_K,H_K; \gamma_K) \}
w_K(H_K; \widehat{\phi_K})
\right]\\
= \int 
\frac{\partial \tilde{Q}_K
}{\partial \gamma_K} \{ W_K - \tilde{Q}_K(A_K,H_K; \gamma_K) \}\\
p(M_K | A_K=a'_k, H_K) p(H_K) dM_K dH_K\\
=\E\left[
\frac{\partial \tilde{Q}_K
}{\partial \gamma_K} \{ \tilde{\E}[W_K | A_K,H_K] - \tilde{Q}_K(A_K,H_K; \gamma_K) \}
\right],
\end{align*}
}
where {\small$\tilde{\E}$} is the expectation taken with respect to the appriopriate conditional distribution derived from {\small$\tilde{p}$}.
Consistency for {\small$\gamma_K$} then follows by standard results on regression estimators.  Given consistency of the stage {\small$i+1$} regression,
we have a consistent estimator for {\small$\tilde{V}_{i+1}$}.  This allows us to repeat the consistency argument for {\small$\tilde{Q}_i$}, as above.
\end{prf}

\begin{thma}{\ref{thm:robust-1-a}}
The estimator in (\ref{eqn:ipw-e}) is consistent and asymptocally normal (CAN) if the models in the set {\small$\{ \pi(W; \psi), p(M | W, A; \phi) \}$} are correctly specified, and  the estimator in (\ref{eq:triple-robust}) is CAN in the \emph{union model}, where any two models in the set
{\small$\{ \pi(W; \psi), \E[Y | A, M, W; \zeta], p(M | W, A; \phi) \}$} are correctly specified.
\end{thma}

\begin{prf}
This proof follows as an extension of consistency results derived for the triply robust estimator of the counterfactual expectation
{\small$\beta = \E[Y(a,M(a'))]$} associated with the natural direct effect in \citep{tchetgen12semi2}.

Assume the models in the set {\small$\{ \pi(W; \psi), p(M | W, A; \phi) \}$} are correctly specified.
Let
{\small
\[
g(W; \psi, \phi) \equiv \frac{
p(M | A=a', W; {\phi})
}{
p(M | A=f(W), W; {\phi})
\widetilde{\pi}(W; {\psi})
}
\]
}
We have
{\scriptsize
\begin{align*}
\E\left[
Y \widetilde{C} g(W; \psi, \phi)
\right]
&=
\E\left[
\E\left[
Y
\widetilde{C} g(W; \psi, \phi)
\middle|
W
\right]
\right]\\
&=
\E\left[
\E\left[
Y | W
\right]
\widetilde{C} g(W; \psi, \phi)
\right].
\end{align*}
}
This is equal to
{\scriptsize
\begin{align*}
& \int Y p(Y | M, A, W) p(M | A,W) p(A | W) \widetilde{C} g(A,W) dA, dM, dY \\
=&
\int Y p(Y | M, A, W) p(M | a',W) p(A | W) \frac{\mathbb{I}(A = f(W))}{ \widetilde{\pi}(W) } dA, dM, dY\\
=&
\int Y p(Y | M, A=f(W), W) p(M | a',W) dM, dY.
\end{align*}
}
This is precisely {\small$\beta$} of interest.


The estimator {\small$\widehat{\beta}_{triple}$} has the form
{\scriptsize
\begin{align*}
& \E\left[
\frac{\widetilde{C}}{\widetilde{\pi}(W; \psi)} \frac{f(M | W, A = a'; \widehat{\phi})}{f(M | W, f(W); \widehat{\phi})} \Big\{ Y -
\mathbb{E}[Y | f(W),M,W; \widehat{\zeta}] \Big\} + \nonumber \right. \\
&
\frac{\mathbb{I}(A=a')}{\pi_a'(W; \widehat{\psi})}
\Big\{ \mathbb{E}[Y | f(W),M,W; \widehat{\zeta}]
- \sum_{M} \mathbb{E}[Y | f(W),M,W; \widehat{\zeta}].\ \nonumber \\
& \left. p(M | W, A = a'; \widehat{\zeta})\Big\} + \sum_{M} \mathbb{E}[Y | f(W), M, W; \widehat{\zeta}] p(M | W, A = a'; \widehat{\phi})
\right],
\nonumber
\end{align*}
}\noindent

Assume {\small$\tilde{\pi}$} was specified incorrectly.
The expectation in the estimator consists of three terms, where the last term is equal to true {\small$\beta$} if models for {\small$Y$} and {\small$M$} are correct.
For the first term we have, by iterated expectation,
{\scriptsize
\begin{align*}
& \E\left[
\widetilde{C} g(W; \phi,\psi) \Big\{ Y -
\mathbb{E}[Y | f(W),M,W; \widehat{\zeta}] \Big\} \right] = \nonumber  \\
& \E\left[
\widetilde{C} g(W; \phi,\psi) \Big\{ \E[Y|A,M,W] -
\mathbb{E}[Y | f(W),M,W; \widehat{\zeta}] \Big\} \right] = \nonumber  \\
& \E\left[
\widetilde{C} g(W; \phi,\psi) \Big\{ \E[Y|f(W),M,W] -
\mathbb{E}[Y | f(W),M,W; \widehat{\zeta}] \Big\} \right] = 0, \nonumber 
\end{align*}
}\noindent
if the {\small$Y$} model is correct.
For the second term we have, by iterated expectation,
{\scriptsize
\begin{align*}
& \E\left[
\frac{\mathbb{I}(A=a')}{\pi_a'(W; \widehat{\psi})}
\Big\{ \mathbb{E}[Y | f(W),M,W; \widehat{\zeta}]
- \sum_{M} \E[Y | f(W),M,W; \widehat{\zeta}] \right. \nonumber \\
& \left. p(M | W, A = a'; \widehat{\zeta})\Big\} \right] =\\
& \E\left[ \E\left[
\frac{\mathbb{I}(A=a')}{\pi_a'(W; \widehat{\psi})}
\Big\{ \mathbb{E}[Y | f(W),M,W; \widehat{\zeta}]
- \sum_{M} \E[Y | f(W),M,W; \widehat{\zeta}] \right. \right. \nonumber \\
& \left. \left. p(M | W, A = a'; \widehat{\zeta})\Big\}\middle| W,A=a' \right] \right] =\\
& \E\left[
\frac{\mathbb{I}(A=a')}{\pi_a'(W; \widehat{\psi})}
\Big\{ \E[ \mathbb{E}[Y | f(W),M,W; \widehat{\zeta}] | A=a',W]\right. \\
&\left. - \sum_{M} \E[Y | f(W),M,W; \widehat{\zeta}]  p(M | W, A = a'; \widehat{\zeta})\Big\} \right] = 0
\end{align*}
}\noindent
if the models for {\small$Y$} and {\small$M$} are correct.

Assume the model for {\small$M$} was specified incorrectly.
The first term in the estimator is mean zero by above argument, since the {\small$Y$} model is still correct.

The second and last terms decompose into
{\scriptsize
\begin{align*}
& \E\left[
\frac{\mathbb{I}(A=a')}{\pi_a'(W; \widehat{\psi})}
\Big\{ \E[ \mathbb{E}[Y | f(W),M,W; \widehat{\zeta}] | A=a',W]\right] \\
& - \E\left[
\frac{\mathbb{I}(A=a')}{\pi_a'(W; \widehat{\psi})}
\sum_{M} \mathbb{E}[Y | f(W),M,W; \widehat{\zeta}]  p(M | W, A = a'; \widehat{\zeta})\Big\} \right] \\
& +\E\left[ \sum_{M} \mathbb{E}[Y | f(W), M, W; \widehat{\zeta}] p(M | W, A = a'; \widehat{\phi}) \right]\\
=&\E\left[
\frac{\mathbb{I}(A=a')}{\pi_a'(W; \widehat{\psi})}
\Big\{ \E[ \mathbb{E}[Y | f(W),M,W; \widehat{\zeta}] | A=a',W]\right]\\
& + \E\left[
\sum_{M} \mathbb{E}[Y | f(W),M,W; \widehat{\zeta}]  p(M | W, A = a'; \widehat{\zeta}) \right.\\
& \left. \E\left[
\left(1 -
\frac{\mathbb{I}(A=a')}{\pi_a'(W; \widehat{\psi})}
\right) \middle| W \right] \right] =\\
& \E\left[
\frac{\mathbb{I}(A=a')}{\pi_a'(W; \widehat{\psi})}
\E[ \mathbb{E}[Y | f(W),M,W; \widehat{\zeta}] | A=a',W]\right] \\
& + \E\left[
\sum_{M} \mathbb{E}[Y | f(W),M,W; \widehat{\zeta}]  p(M | W, A = a'; \widehat{\zeta}) \right.\\
& \left.
\left(1 -
\frac{\pi_a'(W)}{\pi_a'(W; \widehat{\psi})}
\right) 
\right]\\
=& \E\left[
\frac{\mathbb{I}(A=a')}{\pi_a'(W; \widehat{\psi})}
\E[ \mathbb{E}[Y | f(W),M,W; \widehat{\zeta}] | A=a',W]\right]\\
=& \E\left[ \E[ \mathbb{E}[Y | f(W),M,W; \widehat{\zeta}] | A=a',W]\right]
\end{align*}
}\noindent
if the models for {\small$Y$} and {\small$A$} are correct.  The remainder is precisely {\small$\beta$}.

Assume the model for {\small$Y$} was specified incorrectly.
The terms in {\small$\widehat{\beta}_{triple}$} then decompose into
{\scriptsize
\begin{align*}
&\E\left[ \widetilde{C} g(W; \psi, \phi) Y \right]\\
&+ \E\left[ \left( \frac{\mathbb{I}(A=a)}{{\pi}_a(W; \psi)} - \widetilde{C} g(W; \psi, \phi) \right)
	\E[ Y | f(W), M, W; \zeta] \right]\\
& + \E\left[
\sum_{M} \mathbb{E}[Y | f(W),M,W; \widehat{\zeta}]  p(M | W, A = a'; \widehat{\zeta}) \right.\\
& \left. \E\left[ \left(1 - \frac{1 - \widetilde{C}}{1 - \widetilde{\pi}(W;\psi)}\right) \middle| W \right] \right]\\
\end{align*}
}
The last term is mean zero if {\small$\widetilde{\pi}$} is specified correctly.
The second term is equal to
{\scriptsize
\begin{align*}
\int \E[ Y | f(W), M, W; \zeta] p(M | A=a',W) p(W) -\\
\int \E[ Y | f(W), M, W; \zeta] p(M | A=a',W) p(W) = 0
\end{align*}
}
if the {\small$A$} and {\small$M$} models are specified correctly.
The first term is equal to {\small$\beta$} by the argument above.

Both estimators are special cases of the RAL estimator for {\small$\beta$} based on the efficient influence function.  As a result,
standard regularity assumptions \citep{robins92estimating}, and properties of maximum likelihood estimators imply
both estimators are CAN.
\end{prf}

\subsection*{C: Experiments And Visualizations}
\subsubsection*{C1. Models Used In Data Analysis}

We used linear regression with interaction terms between treatment {\small$A_2$} and history {\small$H_2$} to model the outcome {\small$W_2$}: {\small$\E[W_2 | H_2, A_2, {\bf M}_2; \alpha] = \alpha_1 (H_2, A_2, {\bf M}_2) + \alpha_2 A_2 H_2$}, and logistic regression with interaction terms to model all dichotomous variables {\small$X$} with history {\small$H$} and immediate prior treatment {\small$A$}: {\small$logit \big\{p(X = 1 \mid H, A; \beta)\big\} = \beta_1 H + \beta_2 A H$} for {\small$X \in \{{\bf M_1, M_2}, W_1\}$}. 
We used the same form of linear regression with interaction terms to model Q-functions by excluding the mediators: {\small$Q_2(H_2, A_2; \gamma^2) = \gamma^2_1H_2 + \gamma^2_2 A_2 H_2$} and {\small$Q_1(H_1, A_1; \gamma^1) = \gamma^1_1H_1 + \gamma^1_2 A_1 H_1$}. The parameters in all models were estimated by maximizing the likelihood. 

For value search {and G-estimation, we used log CD4 count at the end of sixth month as the outcome of interest, denoted by {\small$W_1$}. We used the same form of models, as described above, for {\small$W_1$}, {\small$\E[W_1 | H_1, A_1, {\bf M_1}; \alpha]$}, and all the mediators, and used logistic regression with no interaction terms to model the treatment assignment: {\small$logit \big\{p(A_1 = 1 \mid H_1; \beta)\big\} = \beta_0 + \beta_1 H_1$}.
We modeled the blip function in (\ref{eq:Gestim-path}) as {\small$\gamma(A_1, H_1; \psi) = \psi_1A_1 + \psi_2 A _1H_1$}. 

\subsubsection*{C2: Decision Tree Visualization Of Learned Policies}
We derived the optimal policies using G-formula, Q-learning, G-estimation, and value search techniques.
The value search method considered a simple class of policies based on a threshold, described in the main body of the paper.
The optimal policies obtained from the first three methods were more complicated functions of prior history.  To aid in interpretability of these policies, we approximated them by means of decision tree multi-label classifiers which treated history as a set of features, and treatment decision as the class label.  The resulting decision tree classifiers are shown in in Fig.~\ref{fig:Gformula}, \ref{fig:Qlearning}, and \ref{fig:g-estimation}.  {In these figures, the label ``path policies'' corresponds to policies that optimize the direct chemical effect of the drug where drug toxicity and adherence behave as if treatment was set to a reference level}. In the following decision trees, the nodes \textit{vl} and \textit{adh} stand for viral load (log scale) and adherence level, repectively. \textit{m00} and \textit{m06} denote the measures at month zero (baseline) and the end of the first six months, and node \textit{who} denotes the stage of disease (there is a total of 4 stages, with higher stages denoting progressively more severe disease). Since classifiers did not achieve perfect accuracy, these decisions trees should be viewed as easy to visualize approximations of the true learned policies. 


{Note that in Fig.~\ref{fig:Gformula}, adherence level is relevant to the overall policy but is omitted in the path specific one. As mentioned above, the classification accuracies are not perfect and hence visualizations are not necessarily a good representation of the true policy. That said, finding the path-specific policy not via {\small$M$}s corresponds to finding the overall policy in the world shown in Fig.~\ref{fig:swig} (b) in the main body. It is true that in this world, adherence at time one, {\small$\tilde{M_1}$}, influences {\small$A_2$}, and as a result it is in principle possible for adherence at time one to be informative in an interesting way for the decision at time two.  However, one large source of variability in patient adherence is precisely due to the treatment we assign, and this source of variability is removed by construction in the world shown in Fig.~\ref{fig:swig}(b) in the main body of the paper -- the world where everyone adheres as if on a reference treatment. A low variability variable is certainly less likely to be relevant for decision making (consider what would happen in the limit where everyone had perfect adherence had they been on a reference treatment). Hence, we are certainly not surprised to find that a path-specific (not via adherence) policy did not include adherence as a relevant variable.}

{\small 
\begin{table}
\centering
\caption{{\small Population log CD4 counts under different policies (under treatment assignments in the observed data, the value is {\small$5.64 \pm 0.01$} in the 2-stage and {\small$5.54 \pm 0.01$} in the 1-stage problem). G-formula and Q-learning are used with 2-stage decision points. Value search and G-estimation are used with 1-stage decision point.}}
\begin{tabular}{c|c} 
 &\small   \shortstack{ \textbf{Path Policies} \\  (not through adherence)}  \\ 
\hline 
\small \textbf{G-formula}      & {\small$ 6.78 \hspace{0.15cm} (5.65, 6.92) $}   \\ 
\small \textbf{Q-learning}     & {\small$ 7.00 \hspace{0.15cm} (4.82, 7.19) $}  
\\ \hline 
\small \textbf{Value search}  &  {\small$ 5.56 \hspace{0.15cm} (5.44, 5.60) $}  \\
\small \textbf{G-estimation}  &  {\small$ 5.56 \hspace{0.15cm} (5.55, 5.58) $}  
\end{tabular}
\label{tab:CD4_adh}
\end{table}
}

\begin{figure*}[t]
\centering
\begin{minipage}[b]{.4\textwidth}
\includegraphics[scale=0.3]{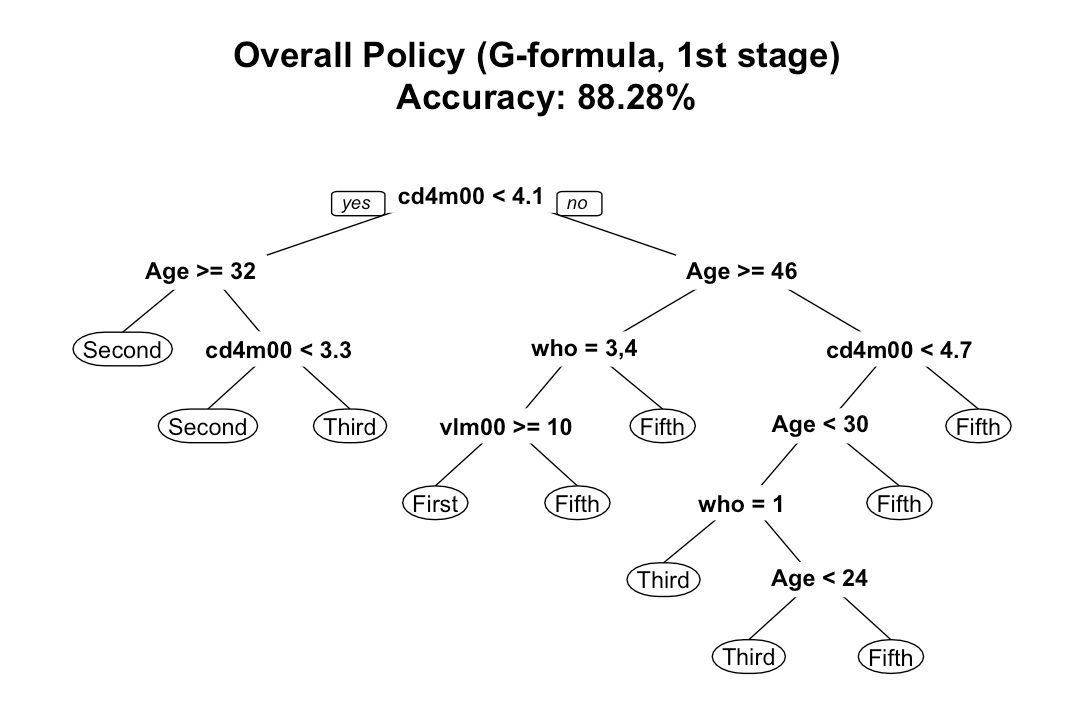}
\end{minipage} \qquad  
\begin{minipage}[b]{.4\textwidth}
\includegraphics[scale=0.3]{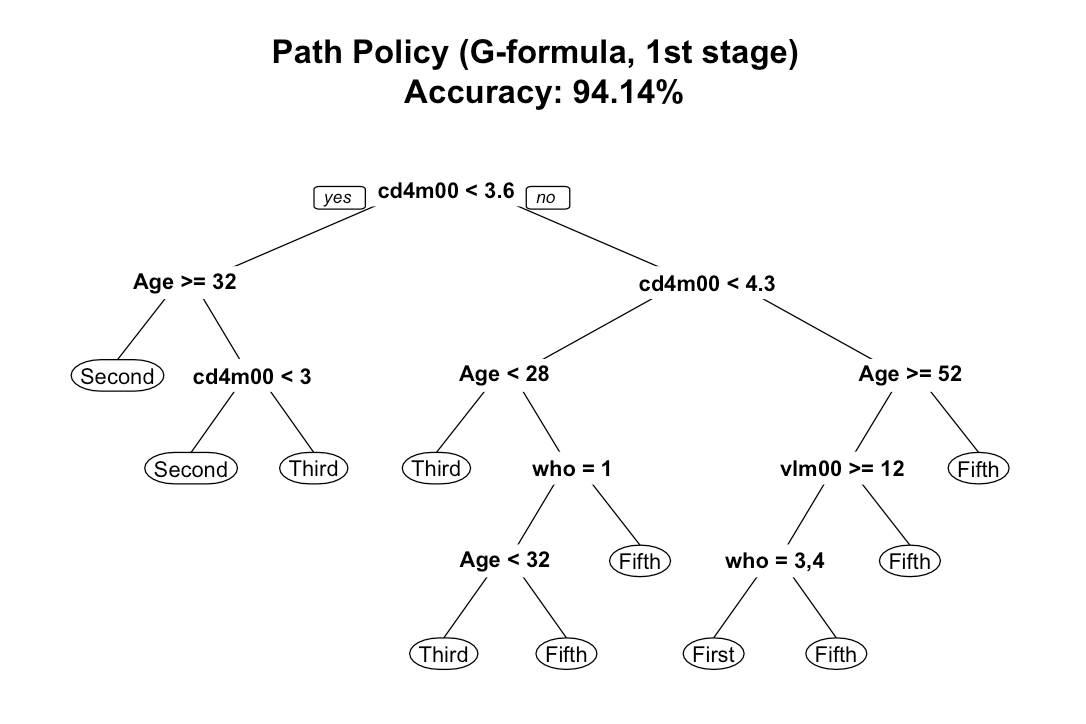}
\end{minipage} 
\hfill
\begin{minipage}[b]{.4\textwidth}
\includegraphics[scale=0.3]{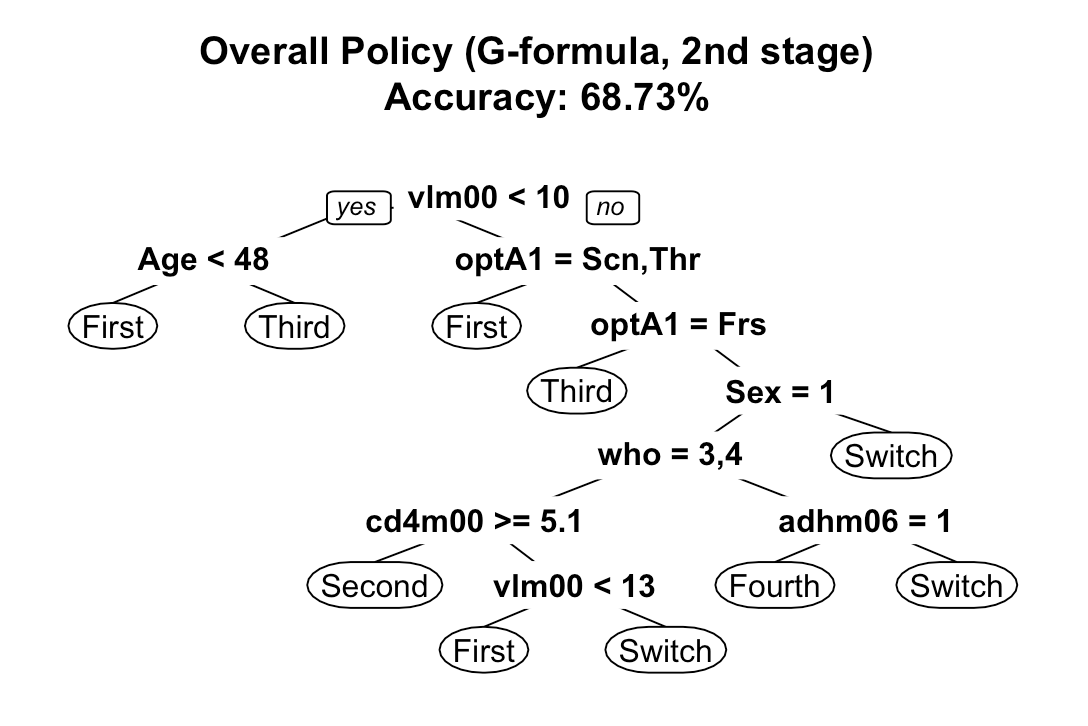}
\end{minipage} \qquad  
\begin{minipage}[b]{.4\textwidth}
\includegraphics[scale=0.3]{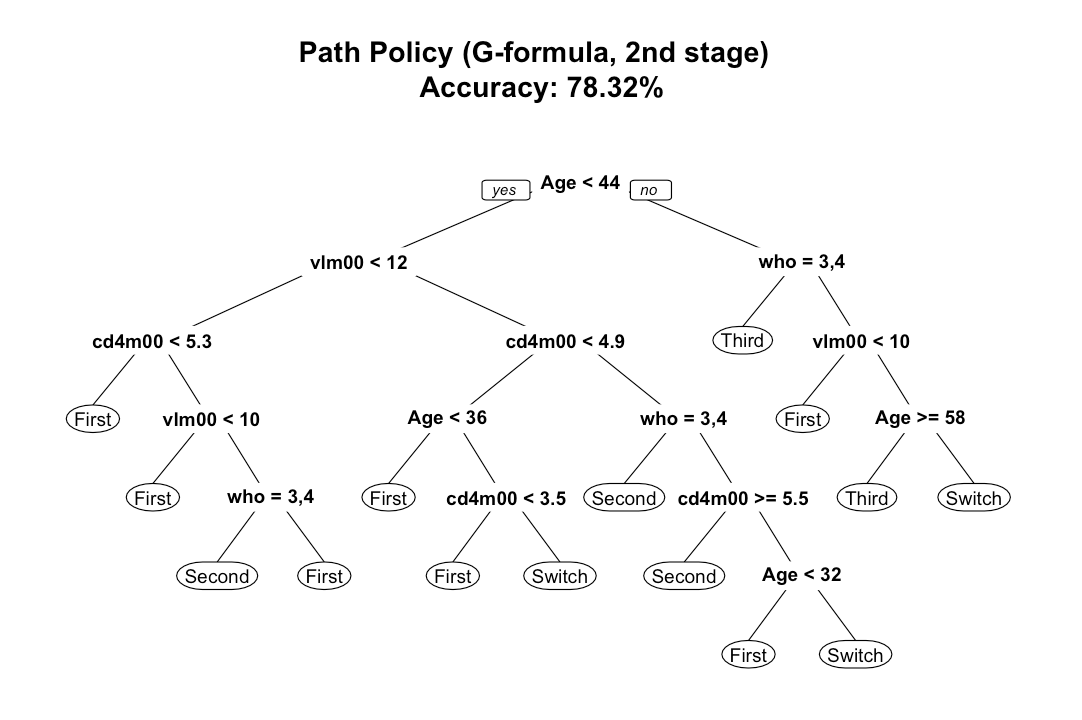}
\end{minipage}
\caption{Decision trees for optimal policies obtained via G-formula. }
\label{fig:Gformula}
\end{figure*}

\begin{figure*}[h]
\centering
\begin{minipage}[b]{.4\textwidth}
\includegraphics[scale=0.31]{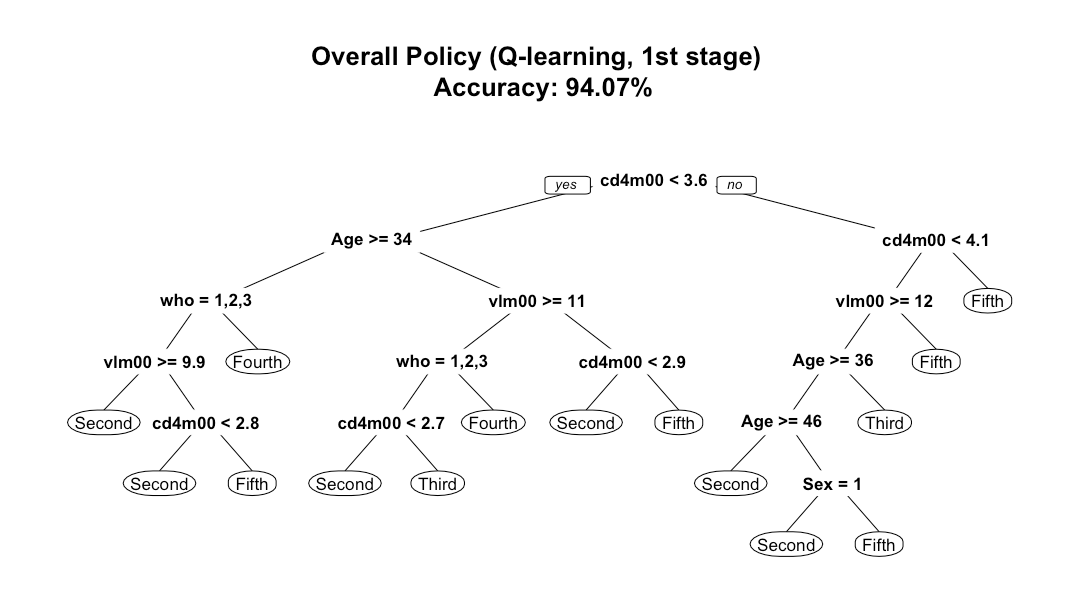}
\end{minipage} \qquad  
\begin{minipage}[b]{.4\textwidth}
\includegraphics[scale=0.31]{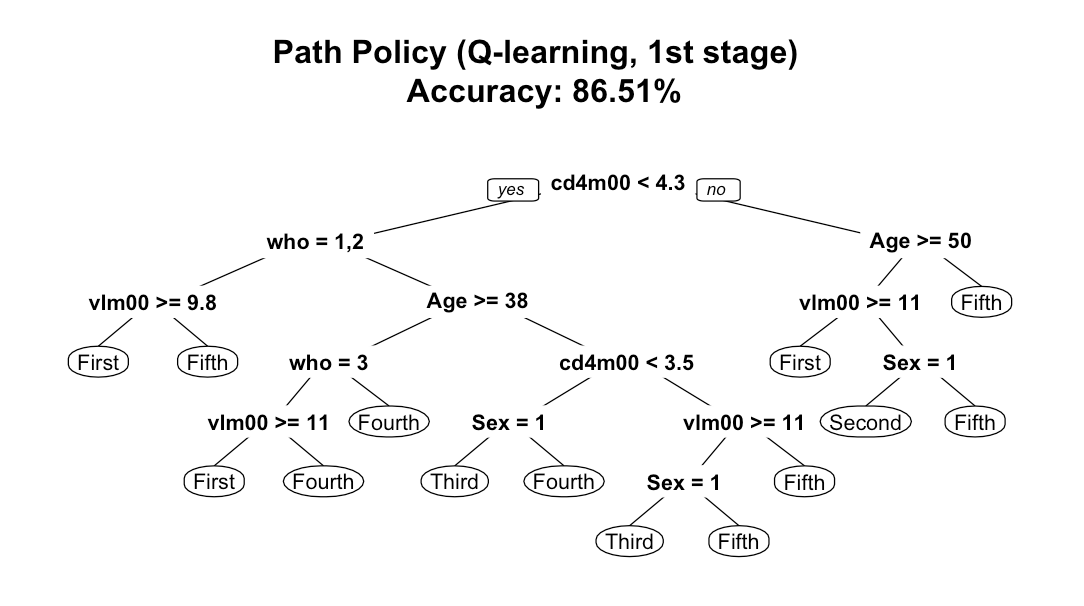}
\end{minipage} 
\hfill
\begin{minipage}[b]{.4\textwidth}
\includegraphics[scale=0.31]{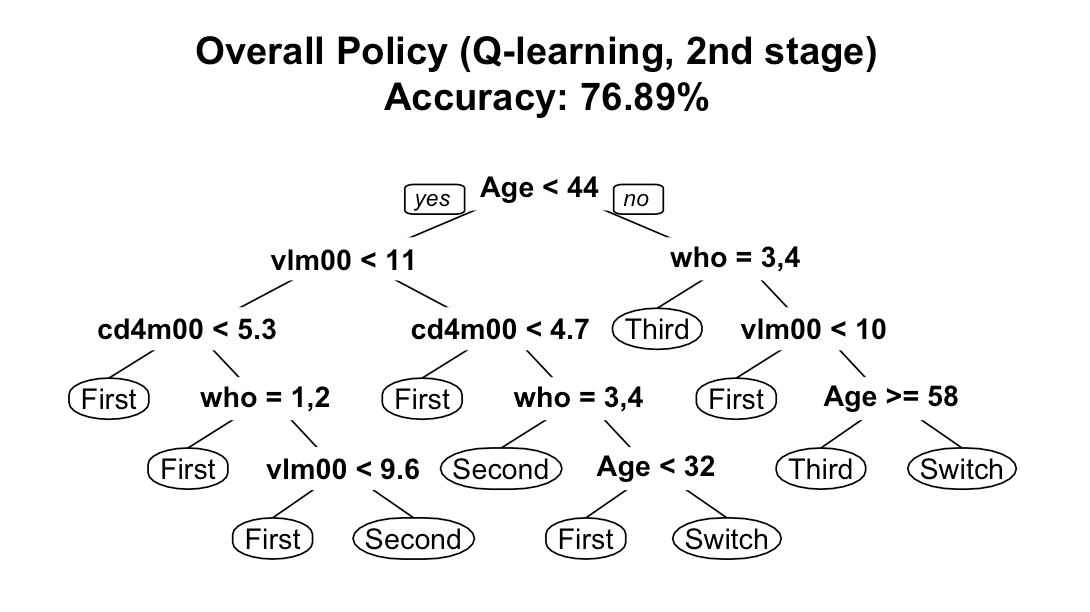}
\end{minipage} \qquad 
\begin{minipage}[b]{.4\textwidth}
\includegraphics[scale=0.31]{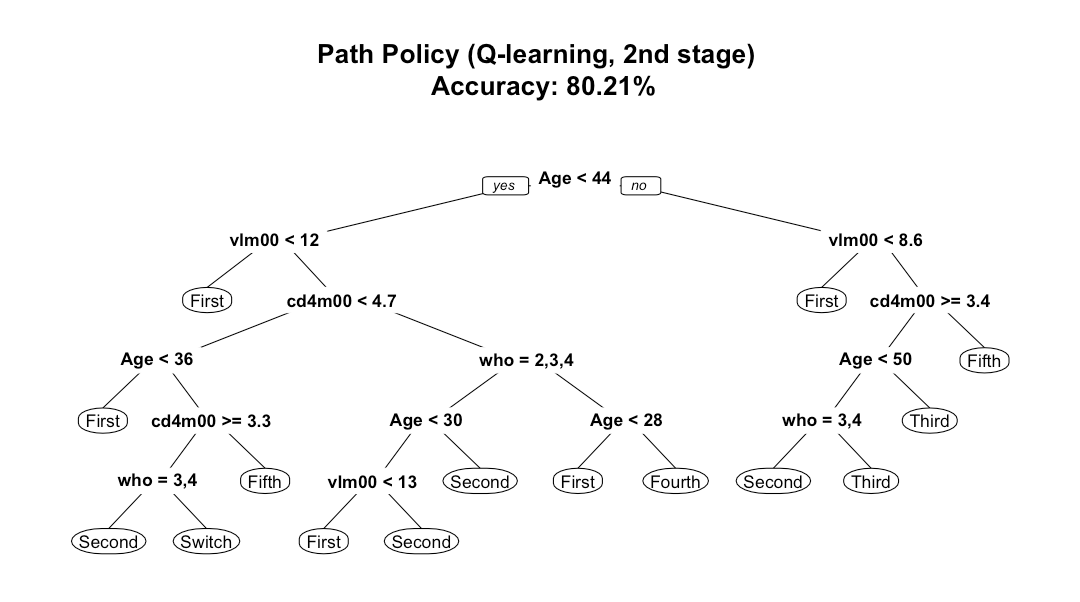}
\end{minipage}
\caption{Decision trees for optimal policies obtained via Q-learning.}
\label{fig:Qlearning}
\end{figure*}

\begin{figure*}
\centering
\includegraphics[scale=0.3]{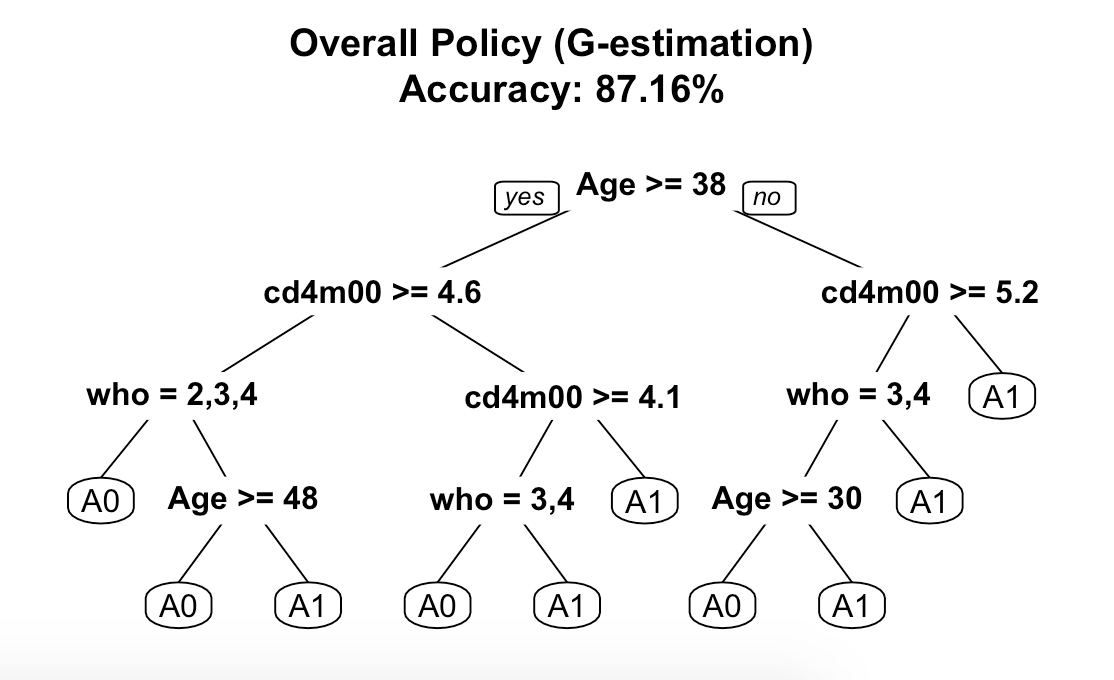} \qquad 
\includegraphics[scale=0.3]{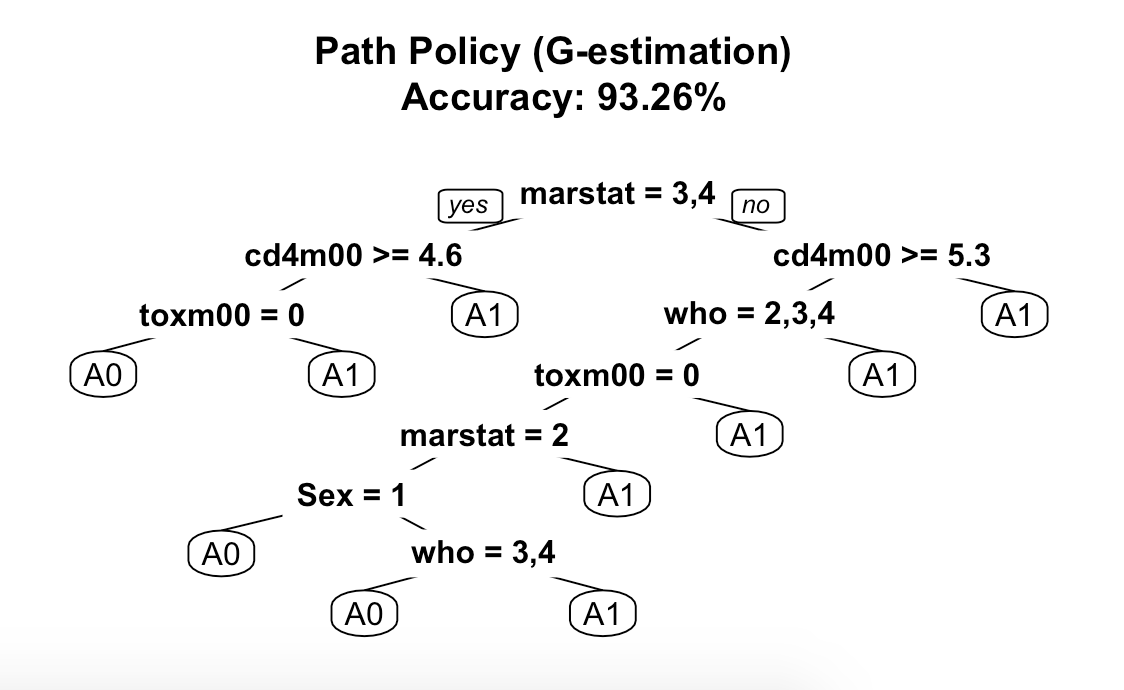} 
\caption{Decision trees for optimal policies obtained via G-estimation. }
\label{fig:g-estimation}
\end{figure*}

\subsubsection*{C3. Additional Experiments}
{To tie the results of this paper to earlier work \cite{miles17quantifying}, we ran additional experiments to find policies that optimize the chemical effect of the drug where only adherence behaves as if the treatments were set to a reference level. Expected outcomes under optimal policies we learned, along with 95\% confidence intervals obtained by bootstrap, are shown in Table.~\ref{tab:CD4_adh}. The results are consistent with the ones  provided in the main body of the paper. For value search, under the same class of policies, {\small$\mathbb{I}\{\text{CD4m00} < \alpha\}$}, and the same modeling assumptions described above, the optimal path policy is chosen to be {\small$\mathbb{I}\{\text{CD4m00} < 550 \text{ cells/mm}^3 \}$}.}


\end{document}